\documentclass[times,twocolumn,final,authoryear]{elsarticle}

\usepackage{ycviu}
\usepackage{framed,multirow}

\usepackage{amssymb}
\usepackage{latexsym}
\usepackage{xcolor}
\usepackage{url}
\usepackage{xcolor}
\definecolor{newcolor}{rgb}{.8,.349,.1}
\usepackage{amssymb}
\usepackage{amsmath,amssymb,amsfonts}
\usepackage{textcomp}
\usepackage{makecell}
\usepackage{xcolor}
\usepackage{tikz}
\usepackage{array}
\usepackage{booktabs}
\usepackage{enumitem}
\usepackage{subcaption}
\usepackage{textcomp}
\usepackage{makecell}
\usepackage{xcolor}
\usepackage{tikz}
\usepackage{array}
\usepackage{booktabs}
\usepackage{enumitem}
\usepackage{newfloat}
\usepackage{listings}
\usepackage{algorithm}
\usepackage{algorithmic}
\usepackage{hyperref}

\newtheorem{remark}{Remark}

\DeclareCaptionLabelFormat{myformat}{\textbf{Fig. #2}}
\usepackage{varioref}
\labelformat{equation}{Eq.~(#1)}
\labelformat{subsection}{Section ~#1}
\labelformat{remark}{Remark ~#1}
\usepackage{float}

\journal{Computer Vision and Image Understanding}
\definecolor{revision}{RGB}{31,33,175}
\begin{document}

\setcounter{page}{1}


\begin{frontmatter}

\title{Large-Scale Riemannian Meta-Optimization via Subspace Adaptation}

\author[1]{Peilin Yu} 
\author[2,1]{Yuwei Wu\corref{cor1}}
\author[1]{Zhi Gao\corref{cor1}}
\cortext[cor1]{Corresponding author.}
\ead{wuyuwei@bit.edu.cn (Yuwei Wu), zhi.gao@bit.edu.cn}
\author[1]{Xiaomeng Fan}
\author[2,1]{Yunde Jia}

\address[1]{Beijing Key Laboratory of Intelligent Information Technology, School of Computer Science \& Technology, Beijing Institute of Technology, China}
\address[2]{Guangdong Laboratory of Machine Perception and Intelligent Computing, Shenzhen MSU-BIT University, China}


\begin{abstract}
Riemannian meta-optimization provides a promising approach to solving non-linear constrained optimization problems, which trains neural networks as optimizers to perform optimization on Riemannian manifolds. However, existing Riemannian meta-optimization methods take up huge memory footprints in large-scale optimization settings, as the learned optimizer can only adapt gradients of a fixed size and thus cannot be shared across different Riemannian parameters. In this paper, we propose an efficient Riemannian meta-optimization method that significantly reduces the memory burden for large-scale optimization via a subspace adaptation scheme. Our method trains neural networks to individually adapt the row and column subspaces of Riemannian gradients, instead of directly adapting the full gradient matrices in existing Riemannian meta-optimization methods. In this case, our learned optimizer can be shared across Riemannian parameters with different sizes. Our method reduces the model memory consumption by six orders of magnitude when optimizing an orthogonal mainstream deep neural network (\emph{e.g.} ResNet50). Experiments on multiple Riemannian tasks show that our method can not only reduce the memory consumption but also improve the performance of Riemannian meta-optimization.
\end{abstract}

\begin{keyword}
Riemannian Meta-optimization\sep Large-scale Optimization \sep Riemannian Manifolds\sep  Subspace Adaptation
\end{keyword}

\end{frontmatter}

\section{Introduction}
\label{sec:intro}
In the computer vision and machine learning communities, many tasks are modeled as optimization problems with non-linear constraints \citep{Huang2018OrthogonalWN,LIU201695,KHAN201497,GARCIASALGUERO2024103887}, which can be efficiently solved by casting the optimization problems on Riemannian manifolds. Riemannian meta-optimization\citep{Gao2020LearningTO} provides a new perspective for optimization on Riemannian manifolds, which trains neural networks as Riemannian optimizers to adapt Riemannian gradients in a data-driven manner. It efficiently reduces human involvement and expert knowledge in employing suitable Riemannian optimizers, showing significant advances across some tasks, such as principal component analysis (PCA) on the Grassmann manifold\citep{yuan2019online}, face recognition on the Stiefel manifold\citep{fan2022efficient} and similarity learning on the SPD manifold\citep{karlinsky2019repmet,Finn2017ModelAgnosticMF}. 

However, existing Riemannian meta-optimization methods take up huge memory footprint for large-scale constrained optimization problems that have a large number of parameters to be optimized (\emph{e.g.}, optimizing a deep neural network on Riemannian manifolds), limiting their applications in practice. The reason behind this issue is that a trained optimizer can only adapt gradients of a fixed size and thus cannot be shared across Riemannian parameters. Existing methods have to train corresponding optimizers for Riemannian parameters of each size, leading to huge memory burdens. For example, considering optimizing a ResNet18 network with orthogonality constraints\citep{Huang2020ControllableOI}, having millions of parameters, the memory consumption of training Riemannian optimizers\citep{9925104} is up to 22.35 GB. The huge memory burdens limit the application of Riemannian meta-optimization.\par
In this paper, we propose an efficient Riemannian meta-optimization method that significantly reduces the memory footprint for large-scale optimization. The core idea of our method is subspace adaptation that trains neural networks to individually adapt row and column subspaces of Riemannian gradients instead of full gradient matrices. Specifically, we compute the row and column covariance matrices of Riemannian gradients and individually feed the diagonal coordinates of the covariance matrices into Long Short Term Memory (LSTM) networks to compute corresponding adaptive weights.
The subspace adaptation scheme allows us to use small LSTM networks that only adapt a single diagonal coordinate each time without considering the size of Riemannian parameters. In this way, the learned optimizer is able to be shared across different Riemannian parameters. Additionally, the covariance matrices contain the second-order information of Riemannian gradients, facilitating the optimization without introducing extra computational loads.

Evaluations on the PCA and face recognition tasks with the Grassmann and Stiefel manifold constraints show that the proposed method learns a better optimizer with much less memory cost than existing Riemannian meta-optimization methods. We further evaluate our method in optimizing orthogonal deep neural networks on the Stiefel manifold and the results demonstrate the effectiveness of our method. For optimizing an orthogonal ResNet50 network, we significantly reduce the memory consumption from 24.34GB to only 40.79KB.

\section{Related Work}

Since the Riemannian stochastic gradient descent (RSGD) algorithm was proposed by Bonnabel\citep{Bonnabel2013StochasticGD}, many works on Riemannian optimization have emerged. Some algorithms generalize accelerated or momentum techniques from Euclidean spaces to Riemannian manifolds to accelerate optimization\citep{liu2017accelerated,Roy2018GeometryAC}, some algorithms focus on reducing the variance of stochastic Riemannian gradients towards faster convergence speed\citep{Zhang2016RiemannianSF,b37}, some others generalize adaptive stochastic gradient algorithms from Euclidean spaces to Riemannian manifolds\citep{Roy2018GeometryAC,Kasai2019RiemannianAS} and some other works focus on studying Riemannian second-order optimization, such as Riemannian Newton-type methods\citep{kasai2018riemannian, hu2018adaptive, si2024riemannian}. All these algorithms fall in the category of designing handcrafted Riemannian optimizers. For example, in RASA\citep{Kasai2019RiemannianAS}, a Riemannian adaptive stochastic gradient algorithm, the rule for generating adaptive weights is designed by hand to maintain matrix
structures of Riemannian parameters during optimization. Compared with the above methods rely on manually designed rules to perform Riemannian optimization, our method leverages Riemannian meta-optimization to automatically learn the optimizer. This allows our optimizer to dynamically adapt step sizes and search directions to task-specific requirements without the need for human involvement
and expert knowledge. In addition, we use the data-driven manner to train the optimizer, through which the optimizer can effectively explore the underlying data distribution and
perform task-specific optimization.

Riemannian meta-optimization methods provide a new perspective for optimization on Riemannian manifolds. Gao et al.\citep{Gao2020LearningTO} first studied Riemannian meta-optimization on SPD manifolds and proposed the Riemannian meta-optimization method (RMO)\citep{9925104} to extend it to various manifolds. They introduced a generalized matrix LSTM model (gmLSTM) to automatically learn the step size and search direction for optimization. Additionally, Fan et al. proposed a gradient-free RMO method that introduces a gradient-free optimizer on tangent spaces\citep{Fan2021LearningAG}, and an implicit differentiation RMO method to efficiently compute the meta-gradient for Riemannian meta-optimization\citep{fan2022efficient}.

The above Riemannian meta-optimization methods utilize the gmLSTM models to parameterize the optimizers. However, for large-scale constrained optimization problems(\emph{e.g.}, optimizing deep neural networks on Riemannian manifolds) when there are a great number of parameters, these methods will consume excessive memory resources. There are two reasons for this: 1) The memory footprint of the gmLSTM model itself is large. Consider optimizing a Riemannian parameter with the size of $d\times p$, the memory cost of storing model parameters of a gmLSTM is up to $O(34d^{2} + 1024dp)$. 
2) The learned optimizer can only adapt gradients of a fixed size and thus cannot be shared across Riemannian parameters with different sizes. These methods have to learn a corresponding optimizer for Riemannian parameters of each size. Hence, in large-scale optimization settings where there are many Riemannian parameters, employing existing Riemannian meta-optimization methods will take up a very large memory footprint. {Our method builds upon existing Riemannian meta-optimization methods by addressing the memory inefficiencies that arise in large-scale optimization settings. In contrast to existing Riemannian meta-optimization methods that adopt the full-matrix adaptation, our method introduces a subspace adaptation scheme. Rather than directly adapting full Riemannian gradients, our optimizer only learns to adapt the row and column subspaces of Riemannian gradients, significantly reducing the dimensionality of the optimization problem. Additionally, our subspace adaptation scheme breaks the dependency on fixed-size gradient updates, allowing the optimizer to be dynamically shared across varying parameter dimensions. Overall, while our method builds on the foundational ideas of Riemannian meta-optimization, it significantly improves the scalability and memory efficiency via subspace adaptation, making it suitable for large-scale and complex Riemannian optimization tasks like optimizing deep neural networks with orthogonality constraints.}

\section{Background}

\subsection{Riemannian Manifold}
\textbf{Notation}. A smooth manifold $\mathcal{M} $ of dimension $d \times p$ can be locally approximated by a Euclidean space $\mathbb{R}^{d \times p}$. Matrix $\boldsymbol{W} \in \mathcal{M}$ denotes a point on manifold $\mathcal{M}$, and the tangent space at $\boldsymbol{W}$ denoted by $T_{\boldsymbol{W} }\mathcal{M} $ contains all vectors that are tangent to $\mathcal{M} $ at $\boldsymbol{W} $.

In order to perform optimization on Riemannian manifolds, the following operations are usually needed.

\textbf{Orthogonal Projection}. Optimization on Riemannian manifolds requires the Riemannian gradient at a point $\boldsymbol{W} \in \mathcal{M} $ to be on the tangent space $T_{\boldsymbol{W} }\mathcal{M} $. The orthogonal projection $\pi _{\boldsymbol{W} }\left ( \cdot \right )$ is used to project a gradient $\nabla _{\boldsymbol{W} }$ in an ambient space onto the tangent space $T_{\boldsymbol{W} }\mathcal{M} $.

\textbf{Retraction}. The retraction operation $\Gamma _{\boldsymbol{W} }\left ( \cdot \right ):T_{\boldsymbol{W} }\mathcal{M} \rightarrow \mathcal{M} $ is used to project an update vector from $T_{\boldsymbol{W} }\mathcal{M} $ to the manifold with a local rigidity condition (details are in Sec. 4.1 of \citep{b42}), and obtain the updated Riemannian parameter on $\mathcal{M} $.

\textbf{Parallel Transport}. $\boldsymbol{W}_{1}$ and $\boldsymbol{W}_{2}$ denote two points on the manifold $\mathcal{M}$. The parallel transport operation $\gamma _{\boldsymbol{W}_{1} \rightarrow \boldsymbol{W}_{2} }\left ( \cdot \right ):T_{\boldsymbol{W}_{1} }\mathcal{M} \rightarrow T_{\boldsymbol{M}_{2}}\mathcal{M} $ transforms a vector from the tangent space $T_{\boldsymbol{W}_{1} }\mathcal{M}$ to the tangent space $T_{\boldsymbol{M}_{2}}\mathcal{M}$ in a ``parallel'' fashion along the geodesic curve connecting $\boldsymbol{W}_{1} $ and $\boldsymbol{W}_{2} $ on $\mathcal{M}$.

\subsection{Riemannian Meta-Optimization}
\label{rmo}
Consider constrained optimization problems, where the goal is to minimize an objective function, 
\begin{equation}
    \begin{aligned}
        \min_{\boldsymbol{W} \in \mathcal{M} }\mathcal{L}\left ( \boldsymbol{W} \right )\triangleq \frac{1}{n}\sum_{i=1}^{n}f\left ( \boldsymbol{W} , \boldsymbol{x}_{i}, \boldsymbol{y}_{i}\right ).
    \end{aligned}
\end{equation}
Here, $\boldsymbol{W} \in \mathcal{M} $ is the parameter of interest, $\left \{ \boldsymbol{x}_{i},\boldsymbol{y}_{i} \right \}_{i=1}^{n}$ denotes the training data and $f(\cdot)$ is the loss function. In Riemannian gradient descent optimization\citep{Roy2018GeometryAC,Bonnabel2013StochasticGD}, the parameter $\boldsymbol{W} $ is updated according to 
\begin{equation}
    \begin{aligned}
	\boldsymbol{W} ^{\left ( t+1 \right )}= \Gamma _{\boldsymbol{W} ^{\left ( t \right )}}\left ( -\boldsymbol{P}^{\left ( t \right )} \right ), 
    \end{aligned}
    \label{W=P}
\end{equation}
where $\boldsymbol{P}^{(t)}$ denotes an update vector on the tangent space at time $t$,
\begin{equation}
    \begin{aligned}
    \boldsymbol{P}^{\left ( t \right )}= \alpha^{\left ( t \right )}\boldsymbol{\xi} ^{\left ( t \right )}.
    \end{aligned}
\end{equation}
Here, $\alpha ^{\left ( t \right )}$ and $\boldsymbol{\xi} ^{\left ( t \right )}$ denote the step size and the search direction at time $t$, respectively. For existing Riemannian optimizers, the step size $\alpha ^{\left ( t \right )}$ and the rule to compute search direction $\boldsymbol{\xi} ^{\left ( t \right )}$ need to be elaborately designed by hand to achieve satisfactory results. 

Instead of designing optimizers by hand, Riemannian meta-optimization methods\citep{9925104, Fan2021LearningAG, fan2022efficient} use the recurrent neural networks, gmLSTM, to adapt the step size $\alpha ^{\left ( t \right )}$ and the search direction $\boldsymbol{\xi} ^{\left ( t \right )}$ by
\begin{equation}
	\begin{aligned}
		\left\{\begin{array}{l}
	\alpha ^{\left ( t \right )}= f_{\phi_{1}} (\boldsymbol{G}^{\left ( t \right )},\boldsymbol{S}^{\left ( t-1 \right )} )\\ 
	
	\boldsymbol{\xi} ^{\left ( t \right )}= \pi _{{\boldsymbol{W}}^{(t)}}\left(g_{\phi_{2}} (\boldsymbol{G}^{\left ( t \right )},\boldsymbol{S}^{\left ( t-1 \right )})\right )
	
\end{array}\right. .
	\end{aligned}
\end{equation}
Here, $\boldsymbol{G}^{\left ( t \right )}$ is the gradient matrix of the loss function at $ \boldsymbol{W} ^{\left ( t \right )}$ and $\boldsymbol{S}^{\left ( t-1 \right )}$ denotes the previous optimization state at time $t-1$. $\phi_{1}$ and $\phi_{2}$ denote the parameters of the functions $f(\cdot)$ and $g(\cdot)$. Concretely, $f_{\phi_{1}}(\cdot)$ and $g_{\phi_{2}}(\cdot)$ are parameterized by the gmLSTM model\citep{9925104}, which uses the architecture of LSTM and replaces the linear operations in LSTM with non-linear Riemannian operations.

\section{Method}

\subsection{Formulation}
Consider a large-scale constrained optimization problem,
\begin{equation}
\begin{aligned}
\min_{\theta}\mathcal{L}&\left ( \theta \right )\triangleq \frac{1}{n}\sum_{i=1}^{n}f(\theta, \boldsymbol{x}_{i}, \boldsymbol{y}_{i})\\
&s.t.\quad \theta=\left \{\boldsymbol{W}_k\in \mathcal{M}_{k} \mid k=1, 2, ..., K\right \} 
\end{aligned} ,
\label{objective ocnn}
\end{equation}
where $\left \{ \boldsymbol{x}_{i},\boldsymbol{y}_{i} \right \}_{i=1}^{n}$ denotes the training data and $f(\cdot)$ is the loss function. Matrix $\boldsymbol{W}_{k}$ denotes the parameter of interest on the $k$-th Riemannian manifold $\mathcal{M} _{k}$, $\boldsymbol{W}_{k} \in \mathbb{R}^{d_{k}\times p_{k}}$ . $\theta$ is composed of all the Riemannian parameters on $K$ manifolds.

Inspired by Shampoo\citep{Gupta2018ShampooPS} and RASA\citep{Kasai2019RiemannianAS}, we utilize two structure-aware matrices $\boldsymbol{R}$ and $\boldsymbol{C}$ to refine the gradient of $\boldsymbol{W}_{k}$. Concretely, for the Riemannian parameter $\boldsymbol{W}_{k}^{\left ( t \right )}$, the structure-aware matrices $\boldsymbol{R}_{k}^{\left ( t \right )}$ and $\boldsymbol{C}_{k}^{\left ( t \right )}$ are used to refine the row and column subspaces of the gradient of $\boldsymbol{W}_{k}^{\left ( t \right )}$, respectively,
\begin{equation}
    \begin{aligned}
\hat{\boldsymbol{G}}_{k}^{\left ( t \right )}=\boldsymbol{R}_{k}^{\left ( t \right )} \boldsymbol{G}_{k}^{\left ( t \right )} \boldsymbol{C}_{k}^{\left ( t \right )}.
    \end{aligned}
    \label{G=RGC}
\end{equation} 
$\boldsymbol{G}_{k}^{\left ( t \right )}$ is the Riemannian gradient matrix of the loss function with respect to $\boldsymbol{W}_{k}^{(t)}$, and $\hat{\boldsymbol{G}}_{k}^{\left ( t \right )}$ is the refined Riemannian gradient. As shown in ~\ref{rmk}, using $\boldsymbol{R}_{k}^{\left ( t \right )}$ and $\boldsymbol{C}_{k}^{\left ( t \right )}$ can preserve the matrix structure of $\boldsymbol{G}_{k}^{\left ( t \right )}$.

\begin{remark}
    \label{rmk}
    Given a gradient matrix $\boldsymbol{G}\in \mathbb{R}^{d\times p}$, and its subspace adaptive matrices $\boldsymbol{R}\in \mathbb{R}^{d\times d}$ and $\boldsymbol{C}\in \mathbb{R}^{p\times p}$, $\hat{\boldsymbol{G}}=\boldsymbol{R}\boldsymbol{G}\boldsymbol{C}$ is able to preserve the matrix structure of $\boldsymbol{G}$. The explanation is as follows. For the matrices $\boldsymbol{G}$, $\boldsymbol{R}$ and  $\boldsymbol{C}$, their vectorial representations satisfy
    \begin{equation}
    \begin{aligned}
       \overline{\operatorname{vec}}(\boldsymbol{R}\boldsymbol{G}\boldsymbol{C})=\left(\boldsymbol{R} \otimes \boldsymbol{C}^{\top}\right) \overline{\operatorname{vec}}(\boldsymbol{G}).
    \end{aligned}
    \label{abc}
    \end{equation}
    The symbol $\otimes$ denotes the Kronecker product, and the $\overline{\operatorname{vec}}$ operator conducts the vectorization
    (or flattening) of a matrix. Here, compared with vectorizing the matrix
    $\boldsymbol{G}$ as $\overline{\operatorname{vec}}(\boldsymbol{G})$, the vectorized $\overline{\operatorname{vec}}(\hat{\boldsymbol{G}})$ (\emph{i.e.} $\overline{\operatorname{vec}}(\boldsymbol{R}\boldsymbol{G}\boldsymbol{C})$) equals multiplying $\overline{\operatorname{vec}}(\boldsymbol{G})$ by a ``structure-preserving'' matrix ($\boldsymbol{R} \otimes\boldsymbol{C}^{\top}$) which is a Kronecker
    product of two matrices containing the subspace information of $\boldsymbol{G}$ to protect its matrix structure. In this way, the refined Riemannian gradient $\hat{\boldsymbol{G}}_{k}^{\left ( t \right )}$ in ~\ref{G=RGC} contains more matrix structure information compared with directly vectorized $\boldsymbol{G}$ which totally destroys the matrix structure.
\end{remark}

Instead of manually designing $\boldsymbol{R}_{k}^{\left ( t \right )}$ and $\boldsymbol{C}_{k}^{\left ( t \right )}$, we propose the subspace adaptation scheme that uses two neural networks $N_{1}(\cdot )$ and $N_{2}(\cdot)$ to individually produce $\boldsymbol{R}_{k}^{\left ( t \right )}$ and $\boldsymbol{C}_{k}^{\left ( t \right )}$,
\begin{equation}
	\begin{aligned}
		\left\{\begin{array}{l}
	\boldsymbol{R}_{k}^{\left ( t \right )}= N_{1}\left (\boldsymbol{G}_{k}^{\left ( t \right )}\right )\\ 
	
	\boldsymbol{C}_{k}^{\left ( t \right )}= N_{2}\left (\boldsymbol{G}_{k}^{\left ( t \right )}\right )
	
\end{array}\right. .
	\end{aligned}
	\label{RC adap}
\end{equation}
In this case, our method produces more suitable adaptation matrices $\boldsymbol{R}_{k}^{\left ( t \right )}$ and $\boldsymbol{C}_{k}^{\left ( t \right )}$ based on the gradient information than manually designing $\boldsymbol{R}_{k}^{\left ( t \right )}$ and $\boldsymbol{C}_{k}^{\left ( t \right )}$. It should be noted that the used neural networks $N_{1}(\cdot )$ and $N_{2}(\cdot)$ can be very small LSTM networks rather than gmLSTMs, which will be elaborated in ~\ref{architecture}. 

Then we use the orthogonal projection $\pi _{\boldsymbol{W}_{k}^{(t)}}(\cdot)$ to project the refined Riemannian gradient $\hat{\boldsymbol{G}}_{k}^{\left ( t \right )}$ back onto the tangent space $T_{\boldsymbol{W}_{k}^{(t)}}\mathcal{M} _{k}$ to obtain the update vector $\boldsymbol{P}_{k}^{\left ( t \right )}$,
\begin{equation}
    \begin{aligned}
    \boldsymbol{P}_{k}^{\left ( t \right )}=\pi _{\boldsymbol{W}_{k}^{(t)}}(\hat{\boldsymbol{G}}_{k}^{\left ( t \right )}),
    \end{aligned}
    \label{P=RGC}
\end{equation}
and use the retraction operation $\Gamma_{\boldsymbol{W}_{k}^{\left ( t \right )}}\left (\cdot \right )$ to update $\boldsymbol{W}_{k}^{\left ( t \right )}$,
\begin{equation}
    \begin{aligned}
    \boldsymbol{W}_{k}^{\left ( t+1 \right )}= \Gamma _{\boldsymbol{W}_{k}^{\left ( t \right )}}\left ( -\boldsymbol{P}_{k}^{\left ( t \right )} \right ).
    \end{aligned}
    \label{w=p}
\end{equation}For ease of reading, we omit the index $k$ in the following sections, unless a clear distinction is necessary.

\begin{figure*}[t]
\begin{minipage}{0.65\textwidth}
\includegraphics[width=1\columnwidth]{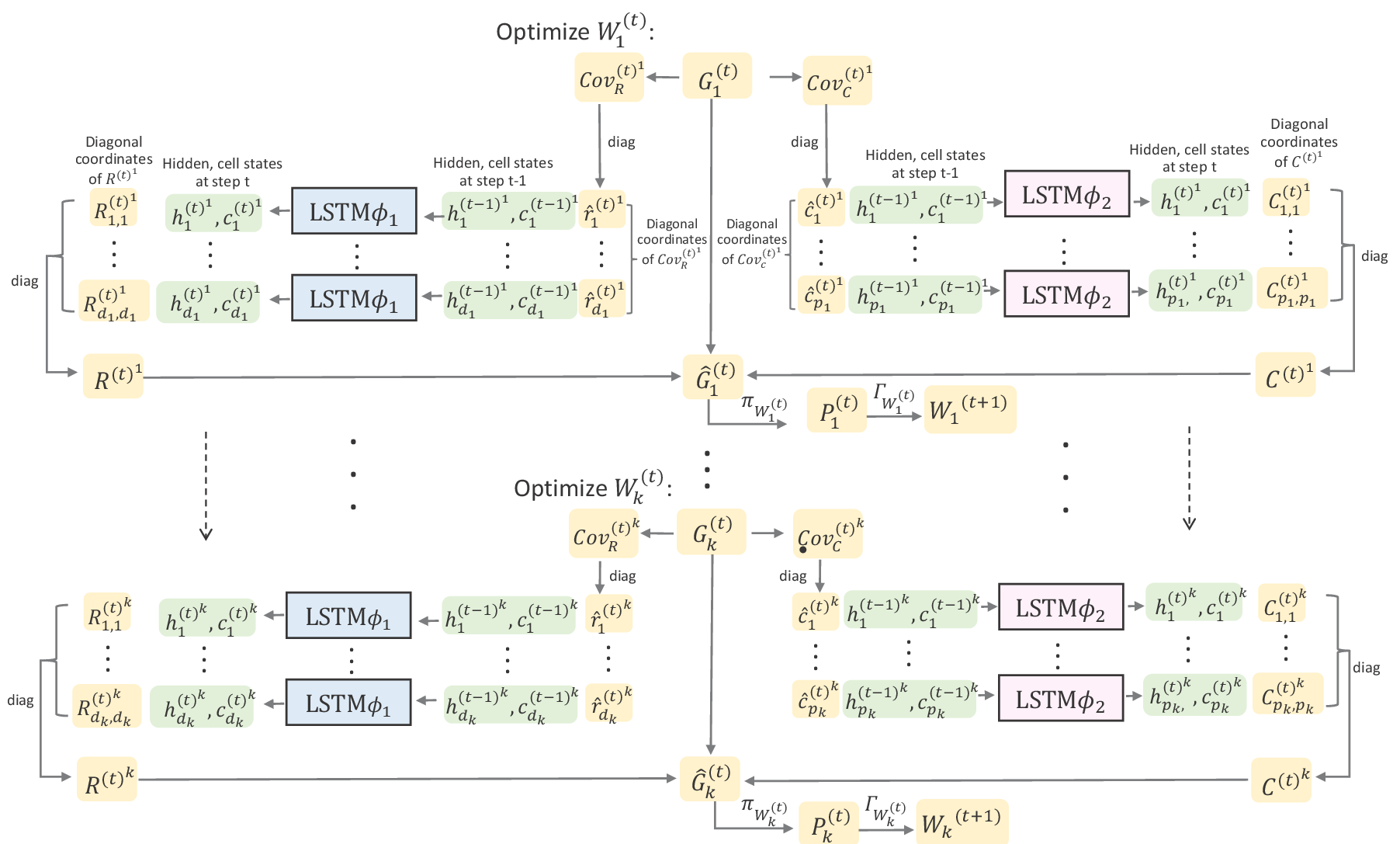}
\caption{The architecture of our optimizer. $\boldsymbol{Cov}_R^{\left ( t \right )}$ and $\boldsymbol{Cov}_C^{\left ( t \right )}$ are the row and column covariance matrices of Riemannian gradient $\boldsymbol{G}^{\left ( t \right )}$. The LSTMs take each coordinate scalar $ \hat{\boldsymbol{r}}_{i}^{\left ( t \right )}$ or $ \hat{\boldsymbol{c}}_{j}^{\left ( t \right )}$, its own previous hidden $\boldsymbol{h}^{\left ( t-1 \right )}$, and cell state $\boldsymbol{c}^{\left ( t-1 \right )}$ as input each time, and respectively produce a diagonal element of adaptive matrix $\boldsymbol{R}^{\left ( t \right )}$ or $\boldsymbol{C}^{\left ( t \right )}$. The LSTM parameters $\phi_{1}$ and  $\phi_{2}$ are shared across coordinates of all Riemannian parameters.}
\label{optimizer arch}
\end{minipage}
\hspace{.03in}
\begin{minipage}{0.325\textwidth} 
\includegraphics[width=1\columnwidth]{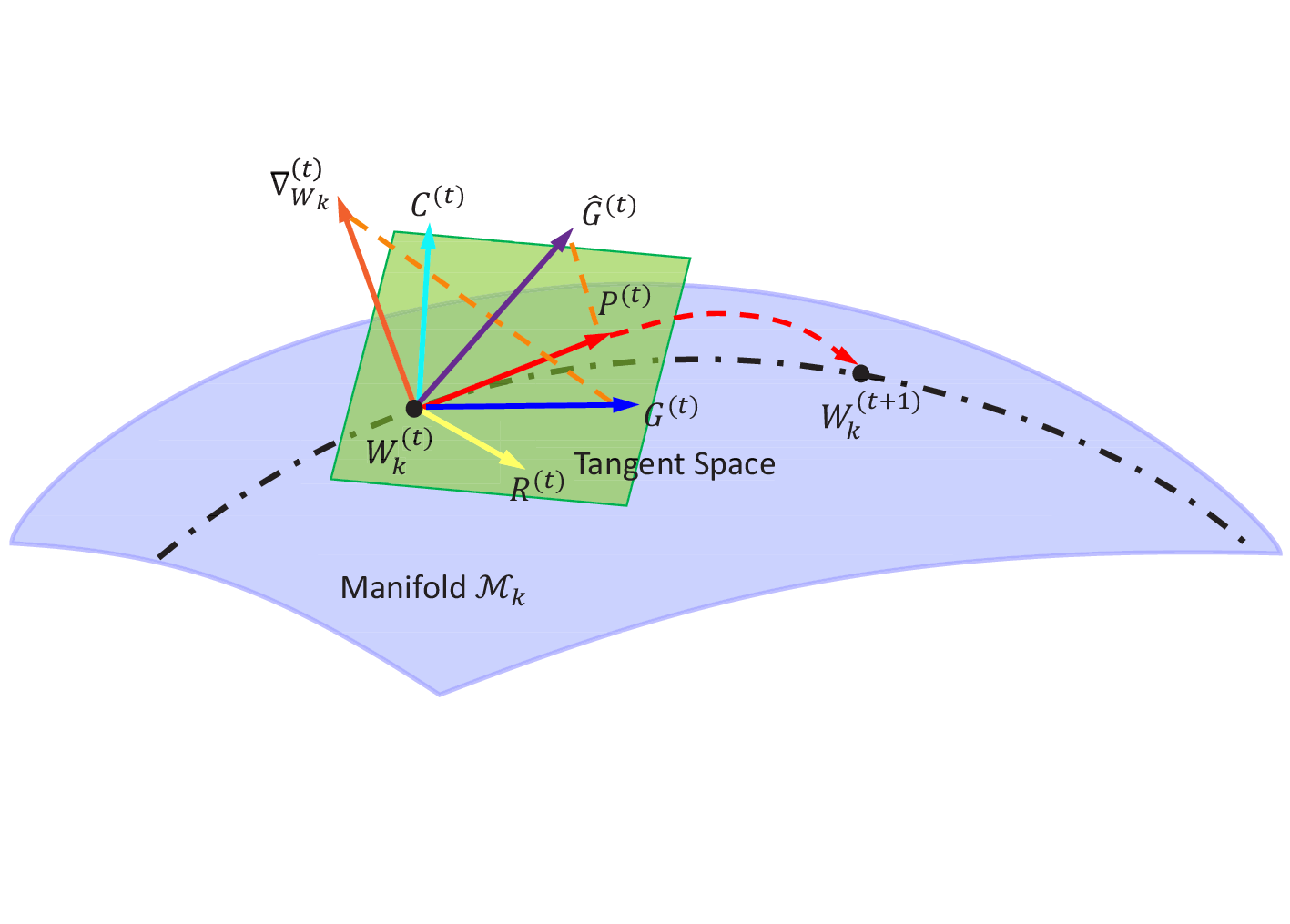} 
\caption{The illustration of the optimization procedure. The black dotted curve denotes a geodesic on manifold $\mathcal{M} _{k}$. The orthogonal projection and retraction operations are denoted as orange and red dotted curves, respectively. $\nabla _{\boldsymbol{W}_{k}}^{(t)}$ (orange solid line) is the Euclidean gradient and $\boldsymbol{G}^{(t)}$ (blue sold line) is the Riemannian gradient on the tangent space. $\boldsymbol{R}^{(t)}$ (yellow solid line) and $\boldsymbol{C}^{(t)}$ (cyan solid line) are the row and column subspace adaptive matrices. $\hat{\boldsymbol{G}}_{k}^{\left ( t \right )}$ (purple solid line) and $\boldsymbol{P}^{(t)}$ (red solid line) are the refined gradient and the update vector, respectively.} 
\label{optimization procedure}
\end{minipage}
\setlength{\abovecaptionskip}{0.0cm}
\end{figure*}

\subsection{Subspace Adaptation}
\label{subsapce adaptsection}
In order to reduce the heavy memory burdens, we introduce the subspace adaptation scheme for Riemannian meta-optimization. Instead of directly adapting the full Riemannian gradient matrix $\boldsymbol{G}^{\left ( t \right )}$ using gmLSTMs, we train small neural networks to individually adapt the row and column subspaces of gradient $\boldsymbol{G}^{\left ( t \right )}$. Here, we first compute the row and column covariance matrices $\boldsymbol{Cov}_R^{\left ( t \right )}$ and $\boldsymbol{Cov}_C^{\left ( t \right )}$ of $\boldsymbol{G}^{\left ( t \right )}$ by
\begin{equation}
	\begin{aligned}
		\left\{\begin{array}{l}
	\boldsymbol{Cov}_R^{\left ( t \right )}=\boldsymbol{G}^{\left ( t \right )}{\boldsymbol{G}^{\left ( t \right )}}^{\top }/p\\ 
	
	\boldsymbol{Cov}_C^{\left ( t \right )}= {\boldsymbol{G}^{\left ( t \right )}}^{\top }\boldsymbol{G}^{\left ( t \right )}/d
	\end{array}\right. .
	\end{aligned}
	\label{transformation}
\end{equation}
$\boldsymbol{Cov}_R^{\left ( t \right )}\in \mathbb{R}^{d\times d}$ and $\boldsymbol{Cov}_C^{\left ( t \right )}\in \mathbb{R}^{p\times p}$ provide the information of the row and column subspaces of $\boldsymbol{G}^{\left ( t \right )}$, and also contain the second-order information of the gradient to facilitate the optimization. 
Then we utilize two recurrent neural networks $N_{1}(\cdot )$ and $N_{2}(\cdot)$ to adapt each subspace, that is, the matrices $\boldsymbol{R}^{\left ( t \right )}$ and $\boldsymbol{C}^{\left ( t \right )}$ in ~\ref{RC adap} are detailed as
\begin{equation}
	\begin{aligned}
		\left\{\begin{array}{l}
        	\boldsymbol{R}^{\left ( t \right )}= N_{1}\left ( {\rm diag}(\boldsymbol{Cov}_R^{\left ( t \right )})\right )\\ 
	
        	\boldsymbol{C}^{\left ( t \right )}= N_{2}\left ({\rm diag}(\boldsymbol{Cov}_C^{\left ( t \right )})\right )
	
        \end{array}\right..
	\end{aligned}
\end{equation}
$\boldsymbol{R}^{\left ( t \right )}\in \mathbb{R}^{d\times d}$ and $\boldsymbol{C}^{\left ( t \right )}\in \mathbb{R}^{p\times p}$ are row and column subspace adaptive matrices, respectively. For computational efficiency, we use the function ${\rm diag}(\cdot)$ that only computes the diagonal vector of a square matrix. 

Then, we feed each coordinate of ${\rm diag}(\boldsymbol{Cov}_R^{\left ( t \right )})$ and ${\rm diag}(\boldsymbol{Cov}_C^{\left ( t \right )})$ into the neural networks $N_{1}(\cdot )$ and $N_{2}(\cdot)$ to learn subspace adaptive matrices $\boldsymbol{R}^{\left ( t \right )}$ and $\boldsymbol{C}^{\left ( t \right )}$.
The coordinatewise operation in ${\rm diag}(\cdot)$ further ensures the learned optimizer to be shared and greatly reduces the memory cost, which will be elaborated in the next section. 

\subsection{Architecture}
\label{architecture}
In our method, we utilize two small LSTM models as $N_{1}\left ( \cdot \right )$ and $N_{2}\left ( \cdot \right )$ to implement the subspace adaptation scheme. The LSTMs take as input the individual diagonal coordinate of $\boldsymbol{Cov}_R^{\left ( t \right )}$ and $\boldsymbol{Cov}_C^{\left ( t \right )}$ with its previous hidden and cell state values in each batch, and produce the adaptive diagonal matrices $\boldsymbol{R}^{\left ( t \right )}\in \mathbb{R}^{d\times d}$ and $\boldsymbol{C}^{\left ( t \right )}\in \mathbb{R}^{p\times p}$. That is, 
\begin{equation}
    \begin{aligned}
        \left\{\begin{array}{l}
	       \boldsymbol{R}_{i,i}^{\left ( t \right )}={\rm LSTM} _{\phi _{1}}\big( \hat{\boldsymbol{r}}_{i}^{\left ( t \right )}, \boldsymbol{h}_{i}^{\left ( t-1 \right )}, \boldsymbol{c}_{i}^{\left ( t-1 \right )}\big )\\ 
	
	        \boldsymbol{C}_{j,j}^{\left ( t \right )}= {\rm LSTM} _{\phi _{2}}\big (\hat{\boldsymbol{c}}_{j}^{\left ( t \right )}, \boldsymbol{h}_{j}^{\left ( t-1 \right )}, \boldsymbol{c}_{j}^{\left ( t-1 \right )} \big )
	
        \end{array}\right.,
    \end{aligned}
    \label{RC_LSTM} 
\end{equation}
where vectors $\hat{\boldsymbol{r}}^{\left ( t \right )}={\rm diag}(\boldsymbol{Cov}_R^{\left ( t \right )})=\left [ \hat{\boldsymbol{r}}_{1}^{\left ( t \right )}, \hat{\boldsymbol{r}}_{2}^{\left ( t \right )}, ..., \hat{\boldsymbol{r}}_{d}^{\left ( t \right )} \right ]$ and $\hat{\boldsymbol{c}}^{\left ( t \right )}={\rm diag}(\boldsymbol{Cov}_C^{\left ( t \right )})=\left [ \hat{\boldsymbol{c}}_{1}^{\left ( t \right )}, \hat{\boldsymbol{c}}_{2}^{\left ( t \right )}, ..., \hat{\boldsymbol{c}}_{p}^{\left ( t \right )} \right ]$ contain the diagonal elements of $\boldsymbol{Cov}_R^{\left ( t \right )}$ and $\boldsymbol{Cov}_C^{\left ( t \right )}$, and vectors $\boldsymbol{h}^{\left ( t-1 \right )}$ and $\boldsymbol{c}^{\left ( t-1 \right )}$ denote their corresponding hidden and cell state values, respectively. 
Compared with gmLSTM, our model is small and only adapts a single coordinate (a scalar) each time to perform optimization. As shown in \autoref{optimizer arch},each coordinate has its own hidden and cell state values but the LSTM parameters are shared across all coordinates. In this case, the learnable optimizer is freed up from being determined by the size of Riemannian parameters and can be shared across different Riemannian parameters. 


In addition, $\boldsymbol{R}^{\left ( t \right )}$ and $\boldsymbol{C}^{\left ( t \right )}$ learned in a data-driven manner can better exploit the geometry of underlying manifold and contain the matrix structure information of the gradient, providing suitable update vector $\boldsymbol{P}^{\left ( t \right )}$ on the tangent space. The optimization procedure is shown in \autoref{optimization procedure} and summarized in \autoref{alg:rsamo algorithm}.

{In our implementation, our LSTM architecture is configured with a default of two layers, each with a hidden state size of 20. The LSTM takes a single diagonal coordinate of the row and column covariance matrices $\boldsymbol{Cov}_R^{\left ( t \right )}$ and $\boldsymbol{Cov}_C^{\left ( t \right )}$, and processes it through the two layers, allowing the network to capture the matrix structure of Riemannian gradient $\boldsymbol{G}$. The output from the LSTM is then passed through two separate linear layers, each corresponding to the row and column subspaces that we adapt in our method.}

\begin{algorithm} [t]
	\caption{Riemannian meta-optimization via subspace adaptation.}
	\label{alg:rsamo algorithm} 
	\textbf{Input}: Initial Riemannian parameters ${\theta }^{\left ( 0 \right )}$ and maximum iteration $T$.
	\begin{algorithmic}[1] 
		\WHILE{ $t \leq  T $ }
			\FOR{each $\boldsymbol{W}_{k}^{\left ( t \right )} \in {\theta}^{\left ( t \right )}$ }
				\STATE
		 		Compute the Riemannian gradient $\boldsymbol{G}_{k}^{\left ( t \right )}=\pi _{\boldsymbol{W}_{k}^{\left ( t \right )}}\left ( \nabla _{\boldsymbol{W}_{k}}^{( t )}\right )$ with respect to the loss in ~\ref{objective ocnn}.
				\STATE
				 Compute the row and column adaptive matrices $\boldsymbol{R}_{k}^{\left ( t \right )}$ and $\boldsymbol{C}_{k}^{\left ( t \right )}$ via subspace adaptation by ~\ref{RC_LSTM} .
				\STATE 
				Compute the update vector $\boldsymbol{P}_{k}^{(t)}$ by ~\ref{P=RGC}.
				\STATE 
				Update $\boldsymbol{W}_{k}^{\left ( t \right )}$ by ~\ref{w=p}.
			\ENDFOR
		\ENDWHILE
  \end{algorithmic}
	\textbf{Output}: Updated Riemannian parameters ${\theta}^{(T+1)}$.
\end{algorithm}

\textbf{Complexity Analysis}. We compare the complexity of adapting a gradient matrix $\boldsymbol{G}\in \mathbb{R}^{d\times p}$ (1) by subspace adaptation using LSTMs and by (2) full-matrix adaptation using gmLSTMs\citep{9925104}. For the both two adaptation schemes, their procedures mainly involves matrix
multiplication and element-wise operations. For a $d\times p$ matrix, the element-wise operations require $\mathcal{O}(dp)$ flops and the matrix multiplication with a $p\times n$ matrix requires $\mathcal{O}(dpn)$. This leads to $\mathcal{O}(16p^{2}d+pd^{2}+18pd)$ flops for using the gmLSTMs to adapt the full gradient matrix $\boldsymbol{G}$. In contrast, the subspace adaptation scheme only adapts $d$ row subspaces and $p$ column subspaces of gradient $\boldsymbol{G}$ and uses the ${\rm diag}(\cdot)$ function that reduces the dimensions of subspace adaptive matrices. In this case, the complexity of subspace adaptation using LSTMs is $\mathcal{O}(16d+16p)$ flops, substantially lower than the full-matrix adaptation scheme.

\subsection{Training}

In our method, we utilize a classic meta-learning framework for training, bi-level optimization (\emph{i.e.}, using an inner loop and an outer loop), which learns the Riemannian parameters and the optimizer in an alternating way. Specifically, the purpose of the inner loop is to optimize the Riemannian parameters $\theta=\left \{\boldsymbol{W}_k\in \mathcal{M}_{k} \mid k=1, 2, ..., K\right \} $ using the learned optimizer, while the outer loop is to optimize the optimizer parameters $\phi=\left \{{\phi _{1}},{\phi _{2}} \right \}$, that is the parameters of LSTMs $N_{1}\left ( \cdot \right )$ and $N_{2}\left ( \cdot \right )$. In the inner loop, Riemannian parameters are updated in $T$ steps by our optimizer with fixed parameter $\phi^{i}$. That is, from $t=i$ to $t=i+T$, $\theta$ is updated with $\phi^{i}=\left \{{\phi _{1}}^{i},{\phi _{2}}^{i} \right \}$ as follows,
\begin{equation}
    \begin{aligned}
        {\theta} ^{\left ( t+1 \right )}=\Gamma _{{\theta} ^{\left ( t \right )}}\left( -\pi_{\theta} ^{\left ( t \right )} \left( N_{{\phi _{1}}^{i}}(\nabla _{\theta} ^{\left ( t \right )})\nabla _{\theta} ^{\left ( t \right )} N_{{\phi _{2}}^{i}}(\nabla _{\theta} ^{\left ( t \right )})\right)\right)
    \end{aligned}.
    \label{inner-update}
\end{equation}

\begin{algorithm} [t]
	\caption{Training.}
	\label{alg:training algorithm} 
	\textbf{Input}: The randomly initialized optimizer, initial Riemannian parameters ${\theta }^{\left ( 0 \right )}$, maximum inner loop iteration $T$ and outer loop iteration $\tau$.
	
	\begin{algorithmic}[1] 
		\WHILE{ $i \leq \tau $ }
		\WHILE{$t \leq T $}
		\STATE
            Compute the inner loss by ~\ref{objective ocnn}.
		\STATE
		Update Riemannian parameters from $\theta^{(t)}$ to $\theta^{(t+1)}$ with ${\phi}^{i}$ by ~\ref{inner-update}.
		\ENDWHILE
		\STATE
		Compute the loss of the optimizer by ~\ref{meta-objective}.
		\STATE
		Update $\phi^{i+1}$ by ~\ref{meta-update}.
		\ENDWHILE
	\end{algorithmic} 
	\textbf{Output}: The learned Riemannian parameters $\theta$ and optimizer with $\phi$.

\end{algorithm} 

In the outer loop, the optimizer with $\phi$ is learned by minimizing the following meta-objective:
\begin{equation}
	\begin{aligned}
		\min_{\phi}\mathcal{J}(\phi)&\triangleq\sum_{t=i}^{i+T} \mathcal{L}\left ( \theta ^{\left ( t+1 \right )} 
		\right )
	\end{aligned},
    \label{meta-objective}
\end{equation}
where $\mathcal{L}\left ( \theta ^{\left ( t+1 \right )} \right )$ is the loss function of the updated Riemannian parameters in the inner loop. In this case, $\phi$ is updated by
\begin{equation}
	\begin{aligned}
		\phi^{i+1}\leftarrow \phi^{i}-\frac{\mathrm{d} \mathcal{J}}{\mathrm{d} \phi}
	\end{aligned} .
    \label{meta-update}
\end{equation}
The training process is summarized in \autoref{alg:training algorithm}.

{By training the optimizer through bi-level optimization in meta-learning, the optimizer learns prior knowledge about how to perform efficient optimization among different tasks. 
Specifically, in meta-learning, the outer-level optimization process guides the model to identify shared patterns across different tasks, which can be leveraged for new tasks. From this, our meta-learned optimizer extracts prior knowledge, that is automatically adjusting step sizes and search directions based on data distributions of given tasks. This allows the optimizer to quickly adapt to new tasks without needing to explore the optimization process from scratch, leading to a good generalization ability to new tasks.}

\section{Experiments}
\subsection{Effectiveness}
\label{effec}

In this section, we evaluate the effectiveness of our method on conventional Riemannian tasks. We compare our optimizer with the existing Riemannian meta-optimization methods: RMO\citep{9925104} and implicit differentiation RMO (\emph{i.e.} I-RMO)\citep{fan2022efficient}, and handcrafted Riemannian optimizers: RSGD\citep{Bonnabel2013StochasticGD}, RSGDM\citep{Roy2018GeometryAC}, RSVRG\citep{Zhang2016RiemannianSF} and RASA\citep{Kasai2019RiemannianAS}. Experiments are conducted on two tasks: principal
component analysis (PCA) on the Grassmann manifold and face recognition on the Stiefel manifold. Following the work RMO and I-RMO, we use the MNIST dataset\citep{lecun1998gradient} for the PCA task and the YaleB dataset\citep{lee2005acquiring} for the face recognition task. The MNIST dataset is a large collection of handwritten digits, and has a training set of 60,000 examples and a test set of 10,000 examples. We resize images to 784-dimensional vectors and aim to learn 128-dimensional (p=128) representations. For face recognition on the Stiefel manifold, we use a linear classifier with the orthogonality constraint. The YaleB dataset contains 2414 images of 38 persons and the training and test sets contain 1900 and 514 images, respectively. For both tasks, we follow the same experimental settings in \citep{fan2022efficient}. 

{In the experiment of principal component analysis (PCA) on the MNIST dataset, the batchsize $n$ is set as 64, the maximum optimization step $\tau$ is set as $\tau=7000$, and the maximum inner iteration is $T=5$. In face recognition on the YaleB dataset, we set $n=64$, $\tau=9000$ and $T=10$. We use the Adam (Kingma, 2015) algorithm with a learning rate of 0.001 to train our optimizer and the optimizer of RMO and I-RMO, with the maximum inner iteration T = 4. The learning rates are set to 0.01 for RSGD, RSGDM, and RSVRG, and 0.0001 for RASA and cRMSProp. The momentum-related $\beta_{1}$ term is set to 0.9 and the $\beta_{2}$ value used in adaptive algorithms is fixed to 0.99 for RSGDM, RSVRG, and cRMSProp.}

\begin{figure}[t]
  \centering
  \begin{subfigure}{0.45\linewidth}
    \includegraphics[width=1.0\columnwidth]{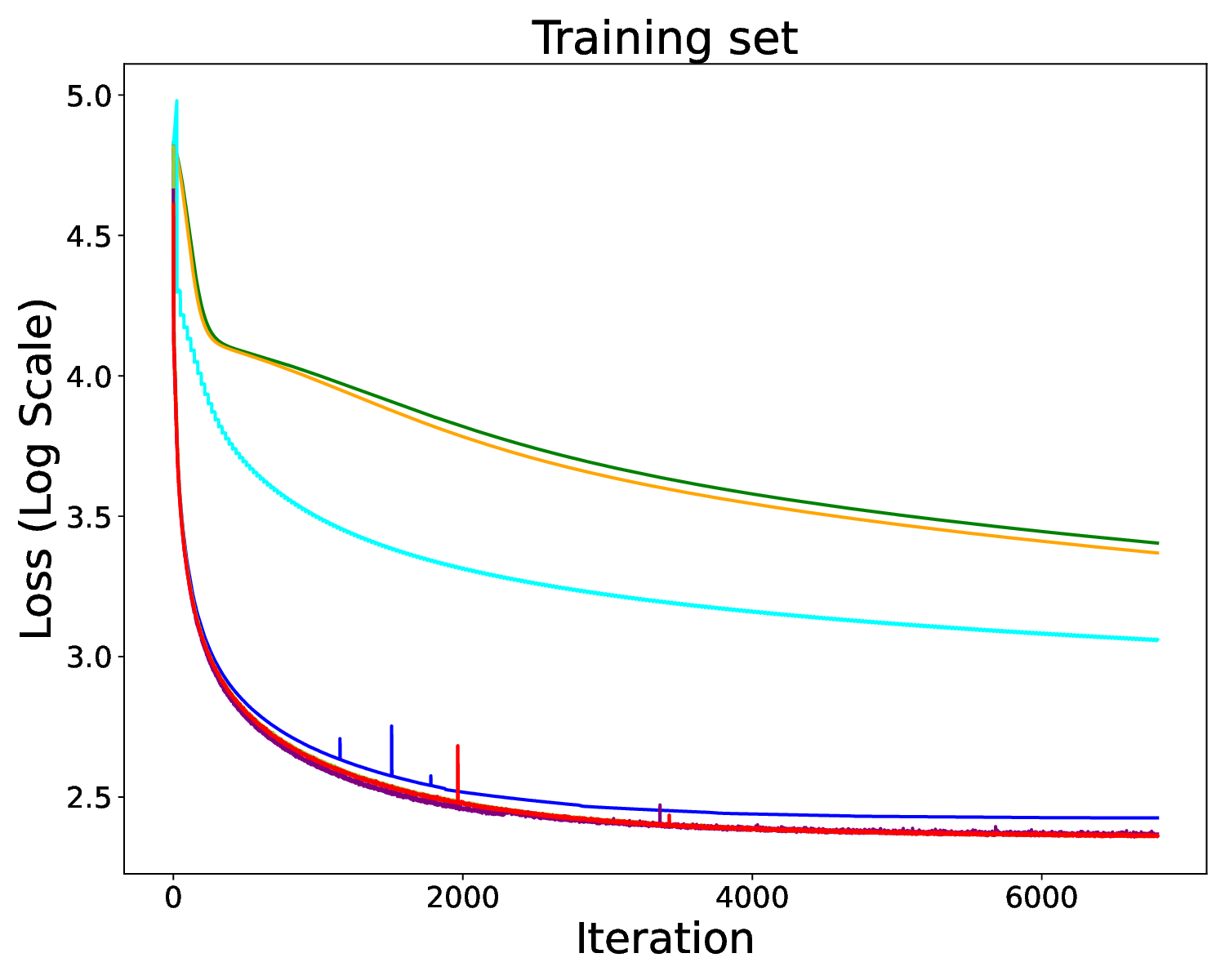}
    \label{fig:pca-train}
    \setlength{\abovecaptionskip}{0.cm}
    \caption{The PCA task}
  \end{subfigure}
  \begin{subfigure}{0.45\linewidth}
    \includegraphics[width=1.0\columnwidth]{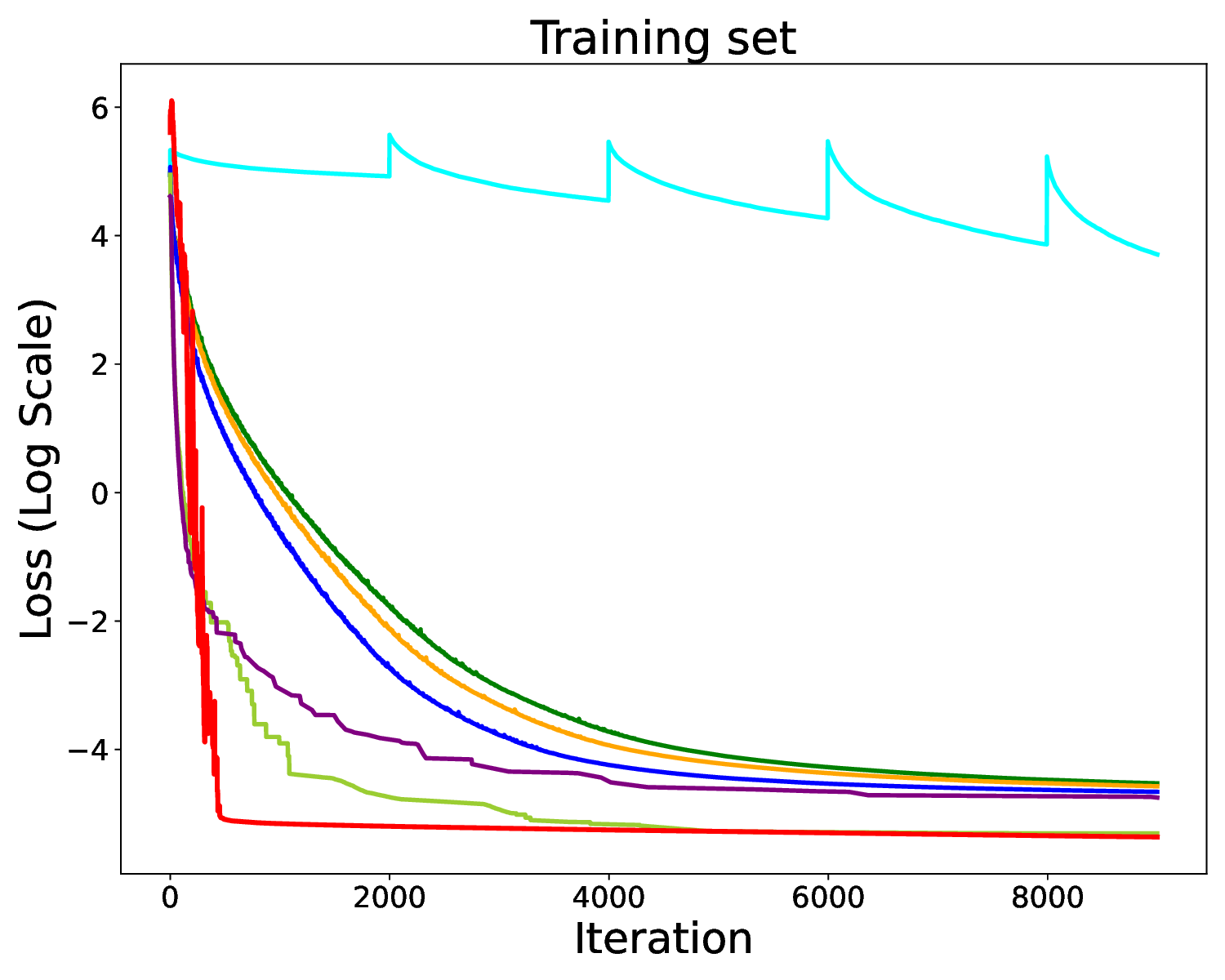}
    \label{fig:face}
    \setlength{\abovecaptionskip}{0.cm}
    \caption{The face recognition task}
  \end{subfigure}
  \begin{subfigure}{0.95\linewidth}
    \includegraphics[width=1\columnwidth]{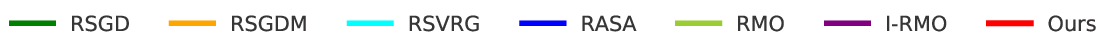}
  \end{subfigure}
  \caption{Plots for the PCA and face recognition tasks (in the log scale).}
  \label{fig:pca_face}

\end{figure}
\begin{table}[t]
    \centering
    \caption{Training time (seconds) and memory usage (MB) of optimizer on the PCA and face recognition tasks.}
    \label{training time}
    \scalebox{0.7}{
    \begin{tabular}{lcc}
        \toprule
         Methods & \makecell[c]{PCA\\Time\hspace{0.9cm} Memory} & \makecell[c]{Face Recognition\\Time\hspace{1cm}Memory}\\
         \hline
     	\specialrule{0em} {1.5pt}{1.5pt}
         RMO\citep{9925104} &
         \makecell[c]{$3.80 \times 10^{-1}$ \hspace{0.1cm} $4.7172\times 10^{2}$ } & \makecell[c]{$1.80 \times 10^{-1}$ \hspace{0.1cm} $2.8800 \times 10^{2}$ }\\
     	\specialrule{0em} {1.5pt}{1.5pt}
         I-RMO\citep{fan2022efficient} & \makecell[c]{$2.90 \times 10^{-1}$ \hspace{0.1cm} $4.7172\times 10^{2}$}& \makecell[c]{$\mathbf{4.00 \times 10^{-2}}$ \hspace{0.1cm} $2.8800 \times 10^{2}$} \\
     	\specialrule{0em} {1.5pt}{1.5pt}
         Ours & \makecell[c]{$\mathbf{1.75 \times 10^{-1}}$ \hspace{0.1cm} $\mathbf{3.9833 \times 10^{-2}}$}& \makecell[c]{$1.73 \times 10^{-1}$ \hspace{0.1cm} $\mathbf{3.9833 \times 10^{-2}}$}\\
         \bottomrule
    \end{tabular}
    }
\end{table}

\subsubsection{Convergence Analysis}
 We analyze the convergence performance of our learned optimizer on the PCA and face recognition tasks. On the PCA and face recognition tasks, we train our optimizer on the training set. \autoref{fig:pca_face} shows that our learned optimizer achieves better performance than hand-designed optimizers, in respect of the convergence speed and the final loss value. This shows the effectiveness of our learned optimizer that captures underlying data structures to obtain a better optimization trajectory. Compared with optimizers learned by RMO and I-RMO, our method performs competitively on the PCA task and clearly surpasses them on the face recognition task.

\subsubsection{Efficiency Analysis}
 We evaluate the training time of each outer loop iteration and the memory usage in training optimizer of our method. We also compare our method with other Riemannian meta-optimization methods RMO and I-RMO, with the inner loop iteration $T=5$. Results on the two tasks are shown in \autoref{training time}. As seen in the table, the memory usage of our optimizer is less than the compared methods RMO and I-RMO on both tasks. The reason is that our method uses the subspace adaptation scheme to learn the optimizer, which trains LSTM networks with a very small memory footprint to individually adapt the row and column subspaces of gradients. RMO and I-RMO both train gmLSTM networks with a large memory footprint to learn the optimizer that needs to adapt the whole gradients. The results demonstrate the efficiency of our method.

\subsection{Application}

\subsubsection{Setting}
We evaluate our method in a typical large-scale constrained optimization problem: optimizing orthogonal neural networks on the Stiefel manifold, which has been widely studied to make efficient and effective training of DNNs possible\citep{FANG2024110281,10056354,Li2021OrthogonalDN,trockman2020orthogonalizing}.

We base the general image classification experiments on three models: VGG16\citep{b49}, ResNet18\citep{b50} and ResNet50\citep{b50}.
We first conduct experiments on three general image classification datasets: CIFAR10, CIFAR100 \citep{b48}, and SVHN (Street View House Numbers)\citep{Netzer2011ReadingDI}. The CIFAR10 and CIFAR100 datasets consist of 60,000 images with a 5-1 training-testing split, divided into 10 and 100 mutually exclusive classes respectively. The SVHN dataset\citep{Netzer2011ReadingDI} is a large-scale dataset that contains a total of 630,420, 32 × 32 colored images, with 73,257 images for training and 26,032 for testing. For ResNet18, we resize input images to $224 \times 224$ pixels, while use the original images ($32 \times 32$ pixels) for ResNet50 considering the computation cost.
We compare our method with the baseline Riemannian meta-optimization methods: RMO\citep{9925104}, I-RMO\citep{fan2022efficient} and handcrafted Riemannian optimizers: RSGD\citep{Bonnabel2013StochasticGD}, RSGDM, cRMSProp\citep{Roy2018GeometryAC}, RSVRG\citep{Zhang2016RiemannianSF} and RASA\citep{Kasai2019RiemannianAS}. We also compare the proposed method with an orthogonalization method in training DNNs: ONI\citep{Huang2020ControllableOI}, and four soft orthogonal regularizers including SO\citep{Xie2017AllYN}, DSO\citep{Bansal2018CanWG}, MC\citep{Bansal2018CanWG} and SRIP\citep{Bansal2018CanWG}.

In the experiments, the associated regularization coefficient $\lambda$ of the regularizers are set to 0.0001 for SO and DSO, and 0.1 for MC and SRIP. We use the Adam\citep{b51} algorithm with a learning rate of 0.001 to train our optimizer and the optimizer of RMO, with the maximum inner iteration $T=4$. For fairness, we use SGD\citep{dekel2012optimal} optimizer with a learning rate of 0.01 and weight decay of 0.0005 to update parameters in the Euclidean space for all algorithms. The learning rates are set to 0.01 for RSGD, RSGDM and RSVRG, and 0.0001 for RASA and cRMSProp. The momentum-related $\beta_{1}$ term is set to 0.9 and the $\beta_{2}$ value used in adaptive algorithms is fixed to 0.99. For all algorithms, we set the batchsize $n$ as 128, and the maximum optimization step $\tau$ as $\tau=12000$ for VGG16 and ResNet18, and $\tau=17000$ for ResNet50 on cifar10. On CIFAR100, we set $\tau=15000$ for VGG16, $\tau=30000$ for ResNet18 and  $\tau=35000$ for ResNet50, respectively. On SVHN, we set $\tau=8000$ for VGG16 and ResNet18, and $\tau=30000$ for ResNet50. 
\par

\begin{figure}[t]
    \centering
    \begin{subfigure}{0.325\linewidth}
    \includegraphics[width=1\columnwidth]{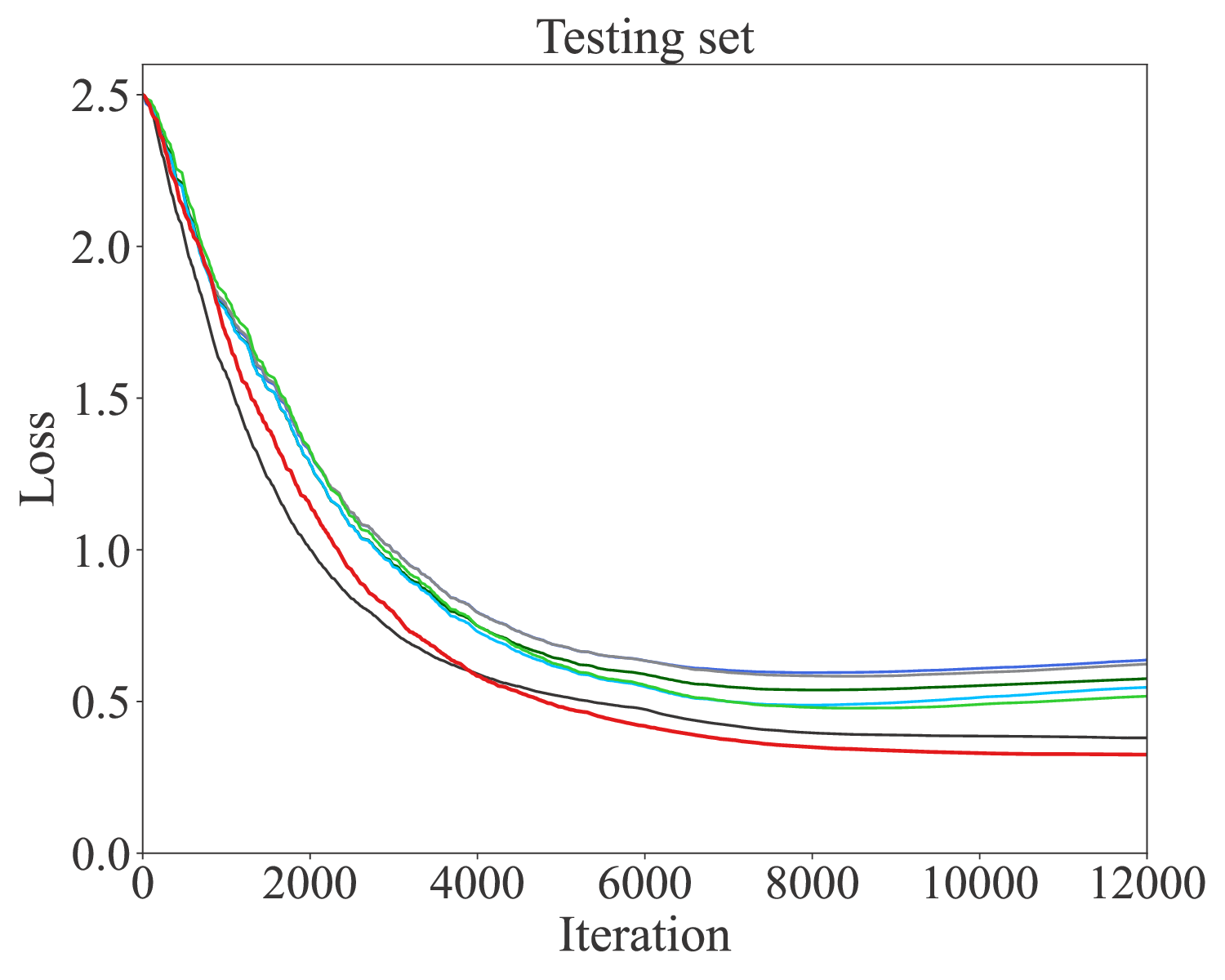}
    \caption{CIFAR10-VGG16}
    \label{fig:cifar10-vgg16}
  \end{subfigure}
  \begin{subfigure}{0.325\linewidth}
    \includegraphics[width=1\columnwidth]{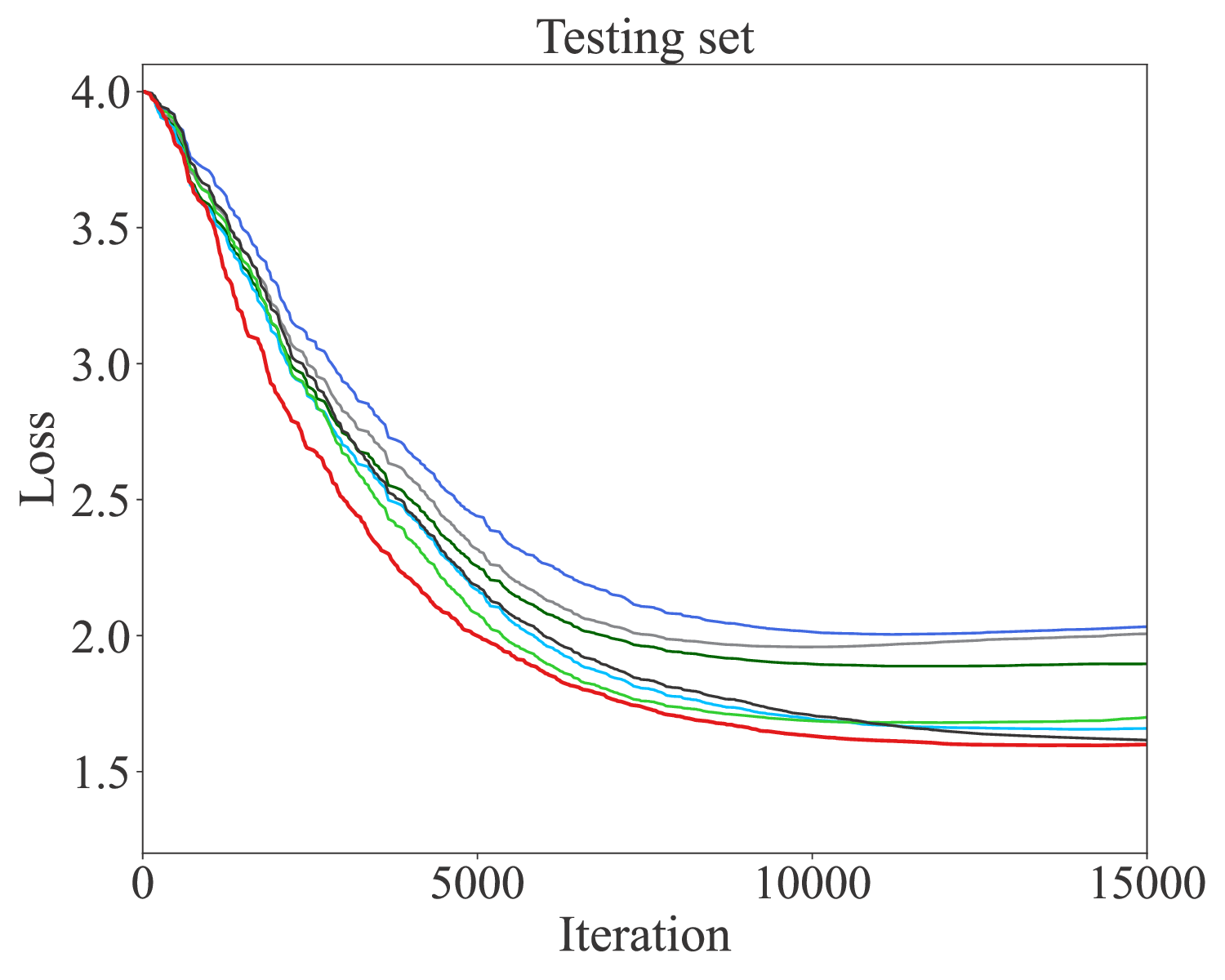}
    \caption{CIFAR100-VGG16}
    \label{fig:cifar100-vgg16}
  \end{subfigure} 
  \begin{subfigure}{0.325\linewidth}
    \includegraphics[width=1\columnwidth]{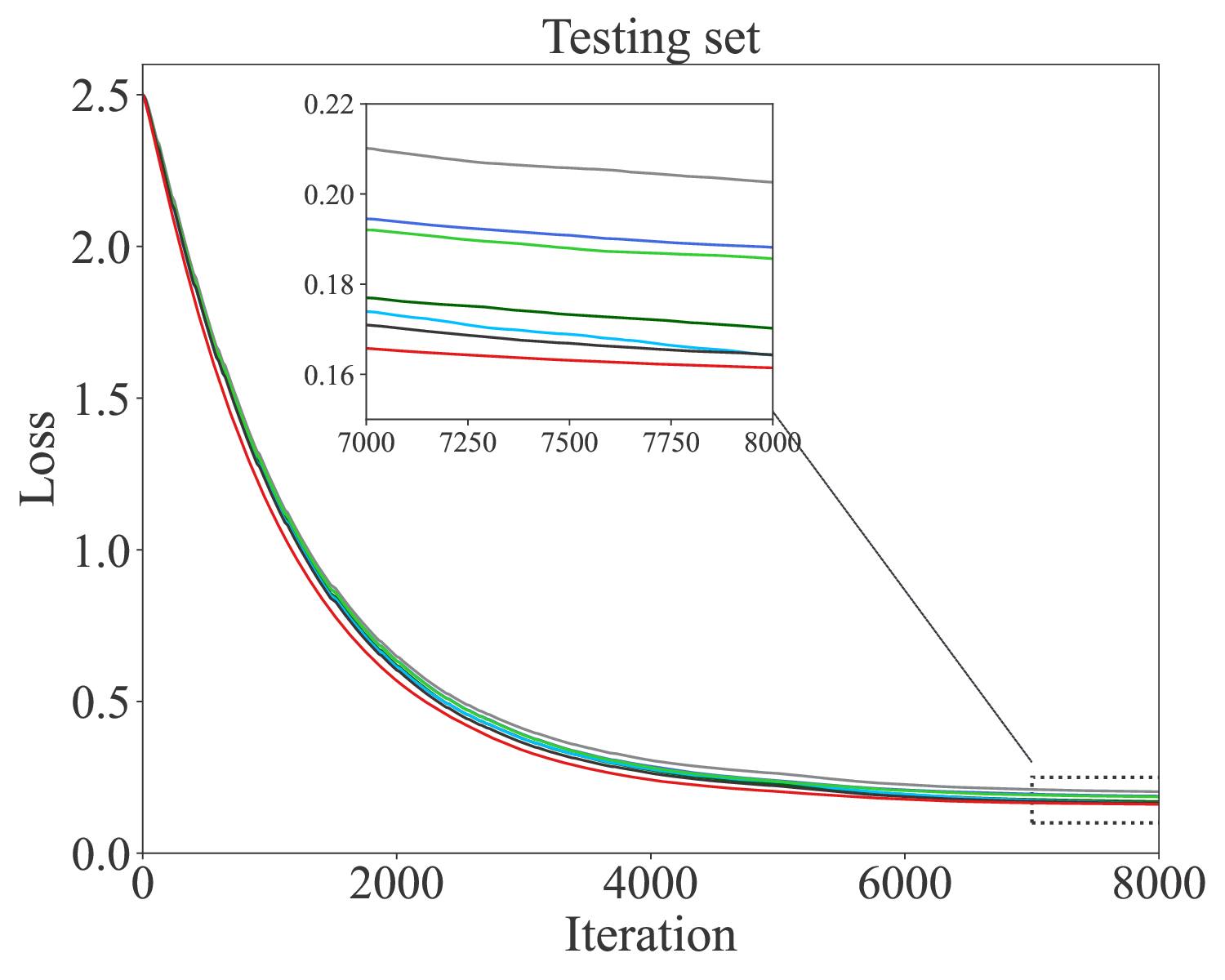}
    \caption{SVHN-VGG16}
    \label{fig:svhn-a}
  \end{subfigure}
  \begin{subfigure}{0.325\linewidth}
    \includegraphics[width=1\columnwidth]{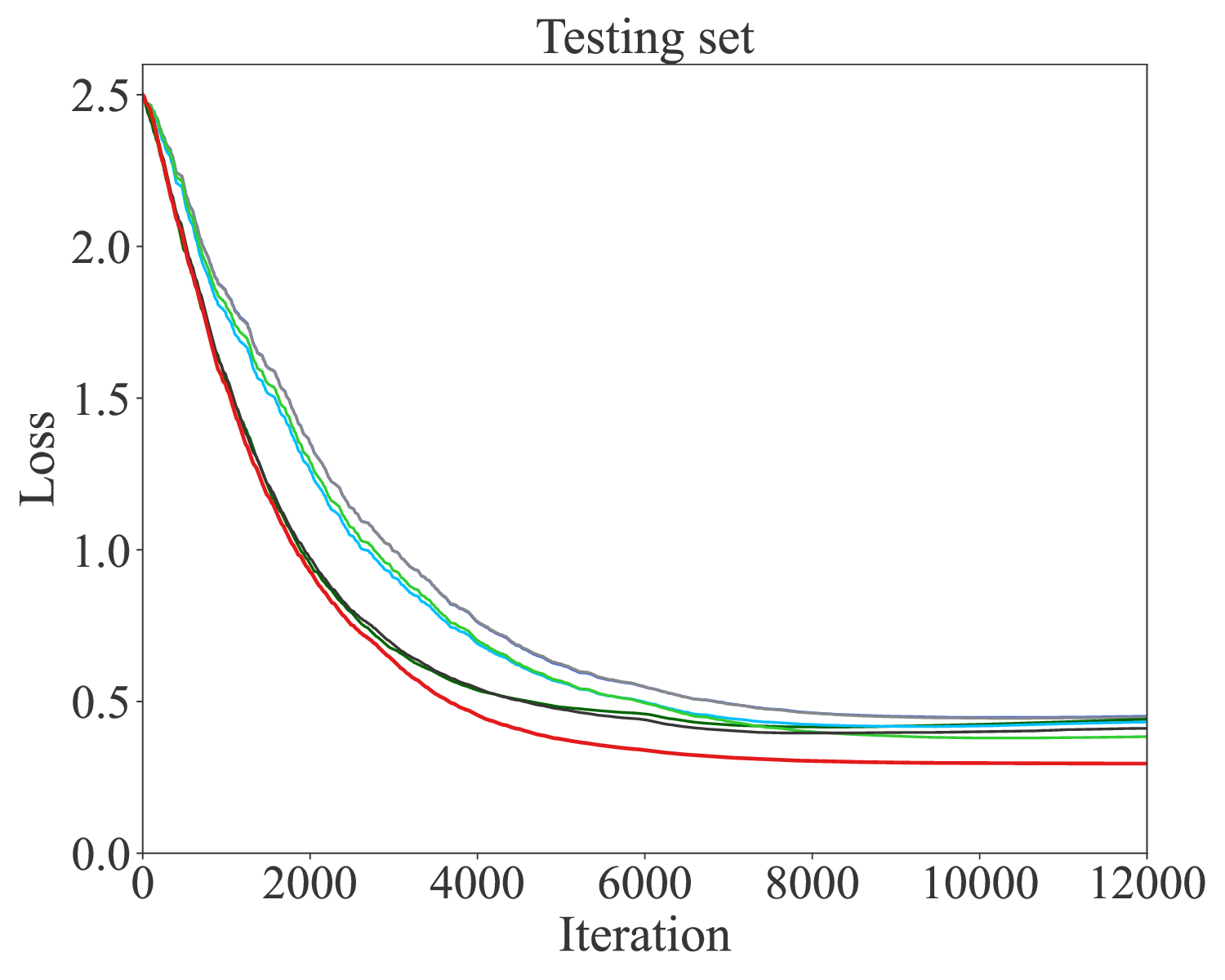}
    \caption{CIFAR10-ResNet18}
    \label{fig:cifar10-resnet18}
  \end{subfigure}
  \begin{subfigure}{0.325\linewidth}
    \includegraphics[width=1\columnwidth]{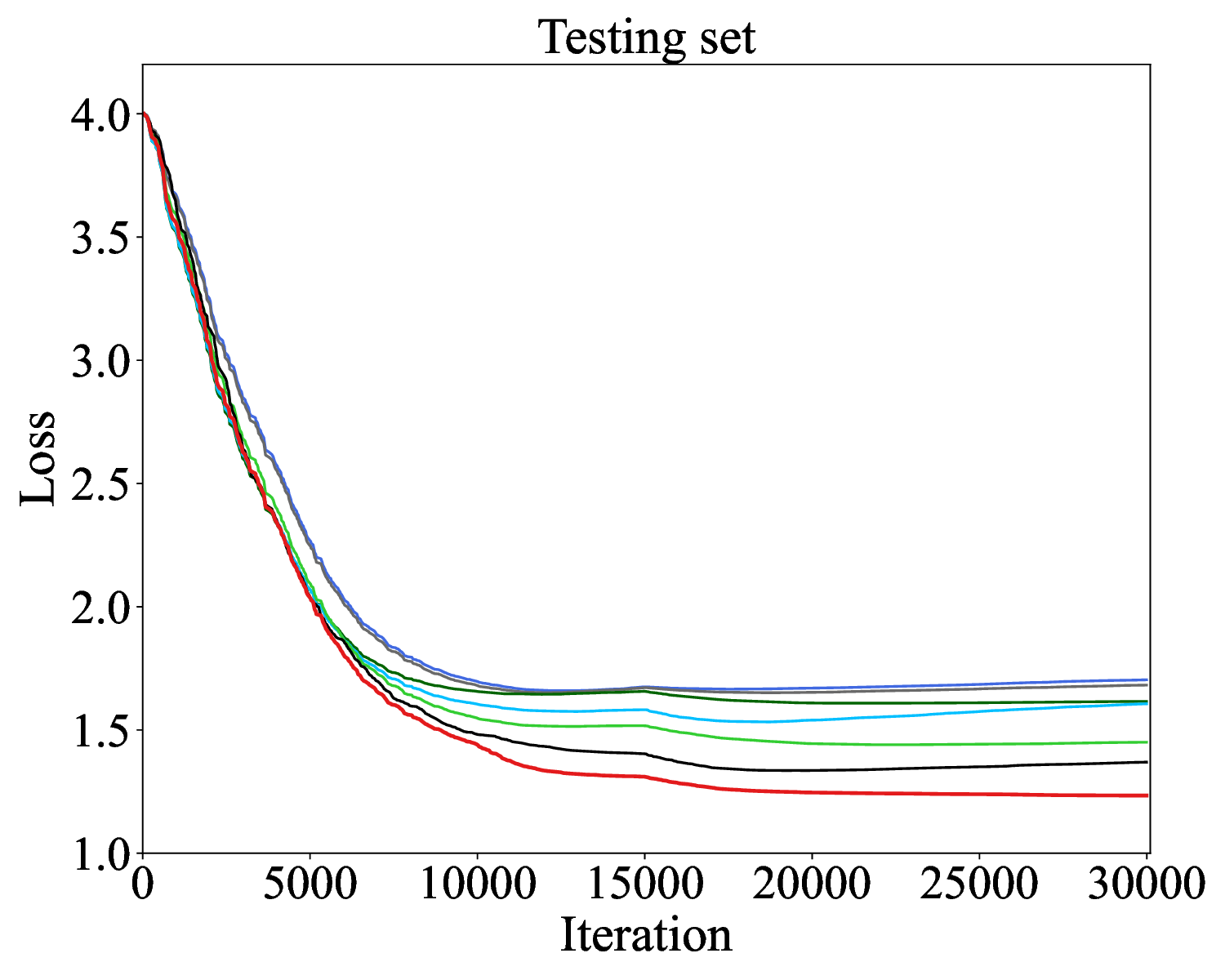}
    \caption{CIFAR100-ResNet18}
    \label{fig:cifar100-resnet18}
  \end{subfigure}
  \begin{subfigure}{0.325\linewidth}
    \includegraphics[width=1\columnwidth]{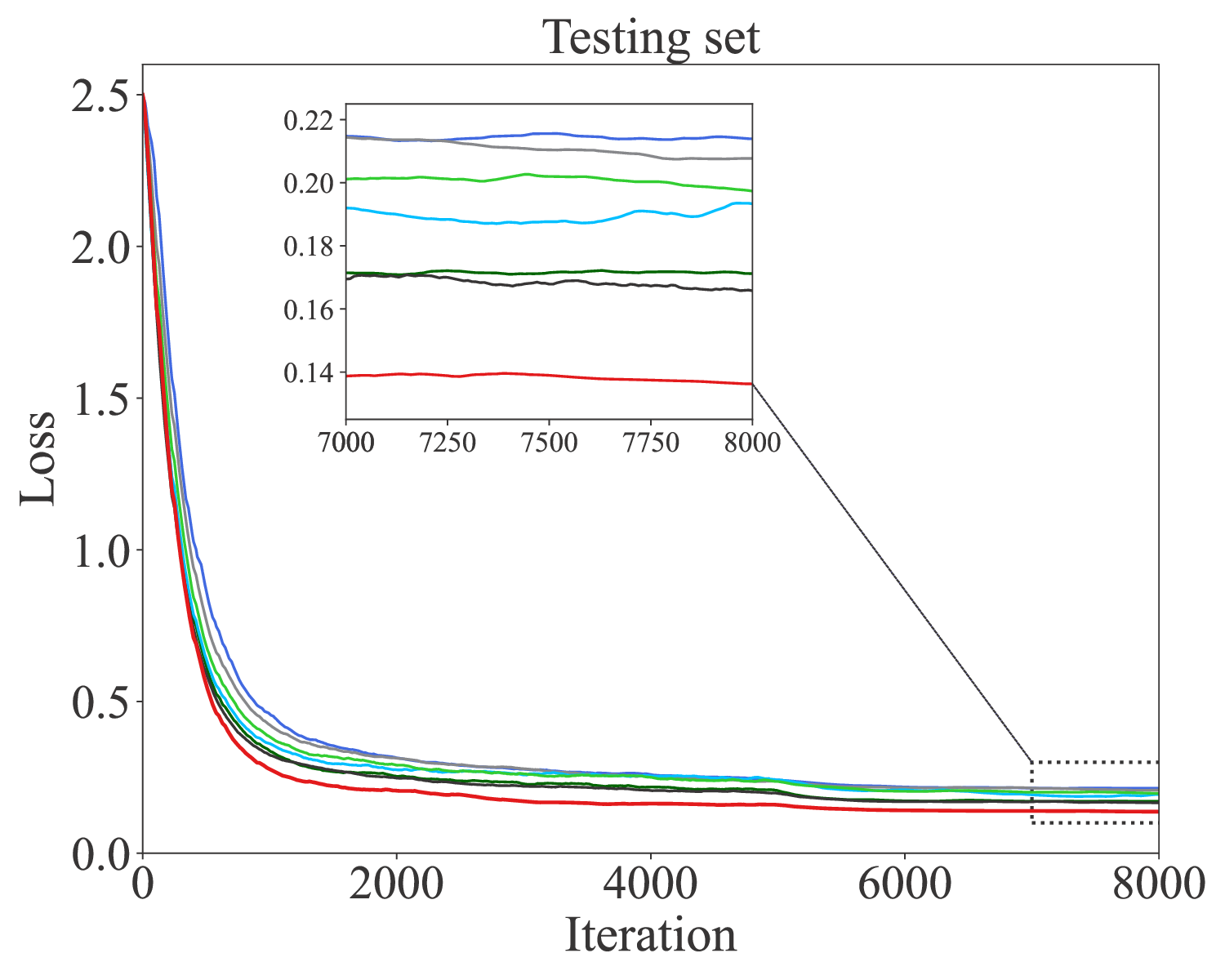}
    \caption{SVHN-ResNet18}
    \label{fig:svhn-b}
  \end{subfigure}
  \begin{subfigure}{0.325\linewidth}
    \includegraphics[width=1\columnwidth]{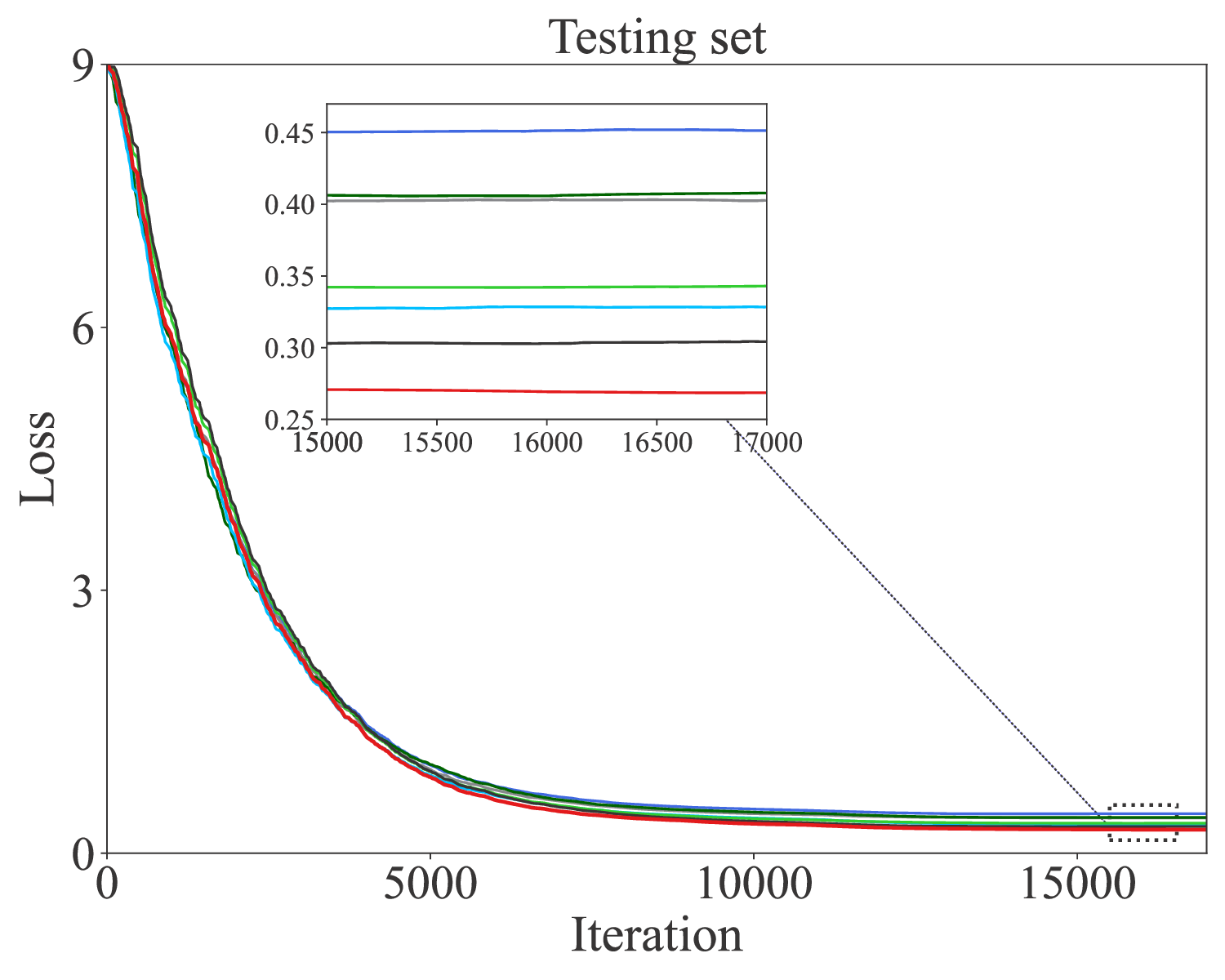}
    \caption{CIFAR10-ResNet50}
    \label{fig:cifar10-resnet50}
  \end{subfigure}
  \begin{subfigure}{0.325\linewidth}
    \includegraphics[width=1\columnwidth]{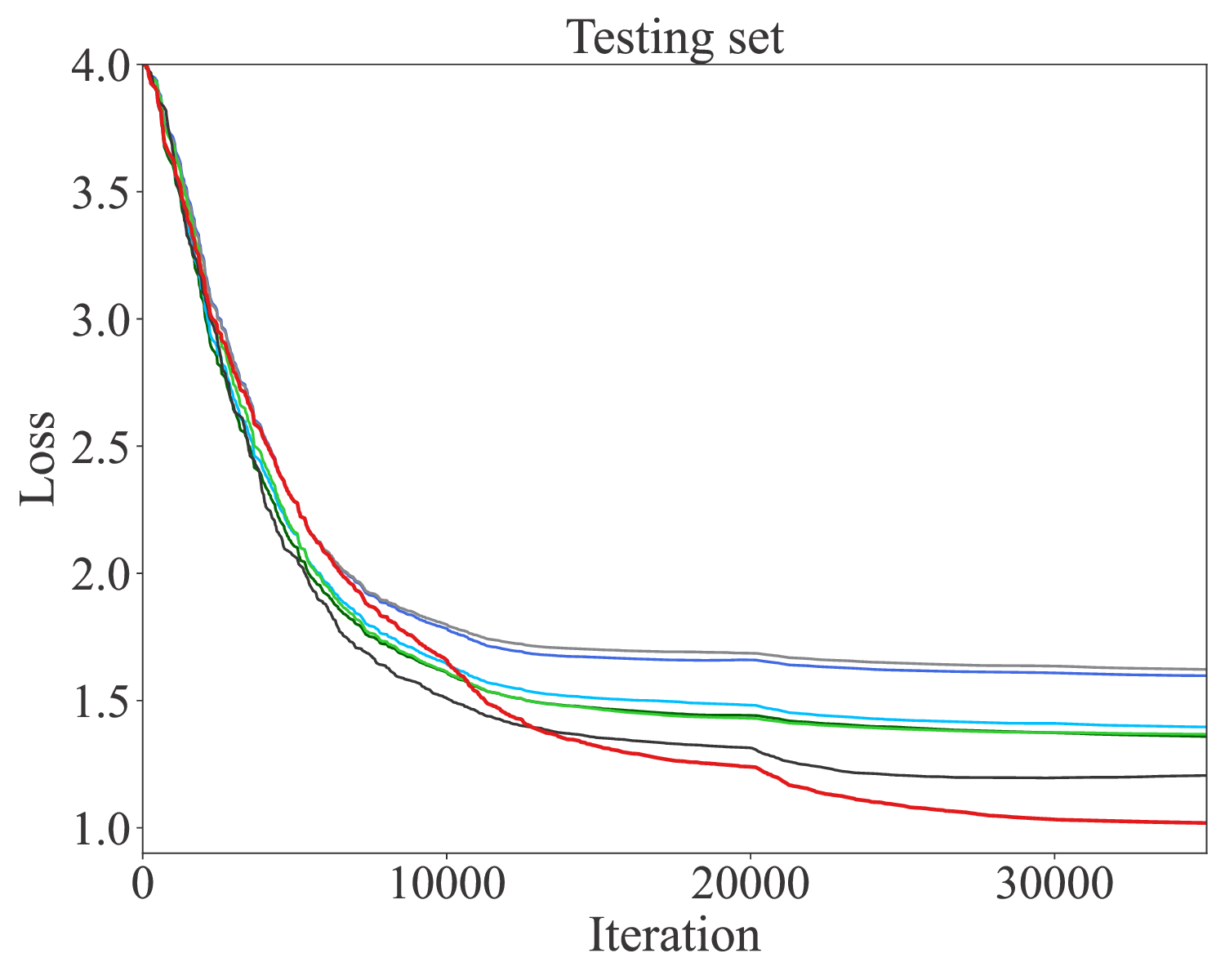}
    \caption{CIFAR100-ResNet50}
    \label{fig:cifar100-resnet50}
  \end{subfigure}
    \begin{subfigure}{0.325\linewidth}
    \includegraphics[width=1\columnwidth]{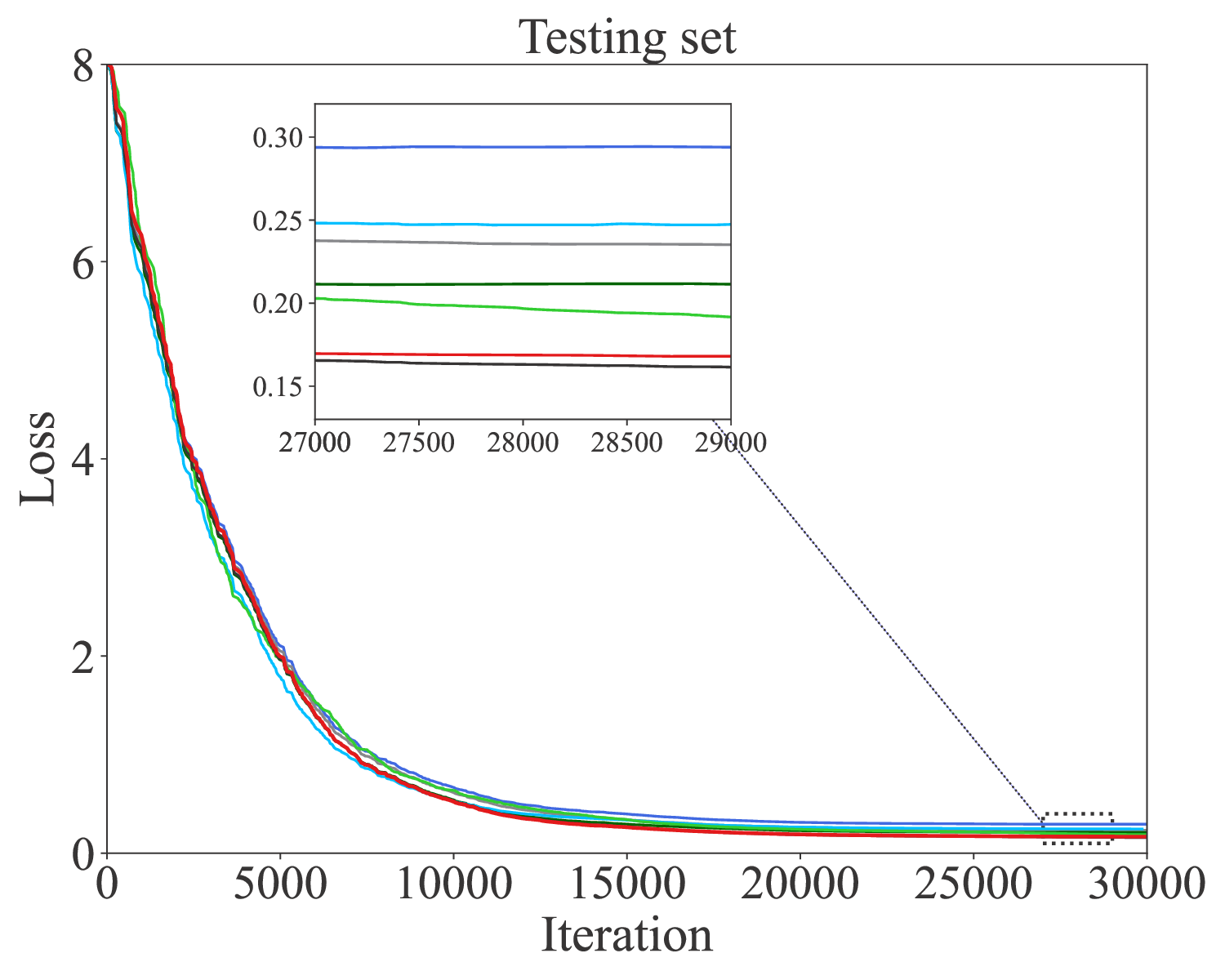}
    \caption{SVHN-ResNet50}
    \label{fig:cifar100-resnet50}
  \end{subfigure}
  \begin{subfigure}{1\linewidth}
    \includegraphics[width=1\columnwidth]{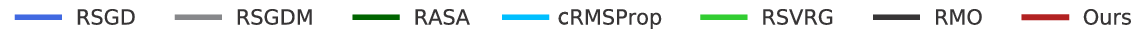}
  \end{subfigure}

  \caption{Testing loss curves of different Riemannian optimization algorithms on optimizing VGG16, ResNet18 and ResNet50.}
  \label{loss curve resnet}
\end{figure}
\begin{figure}[t]
  \centering
  \begin{subfigure}{0.325\linewidth}
    \includegraphics[width=1\columnwidth]{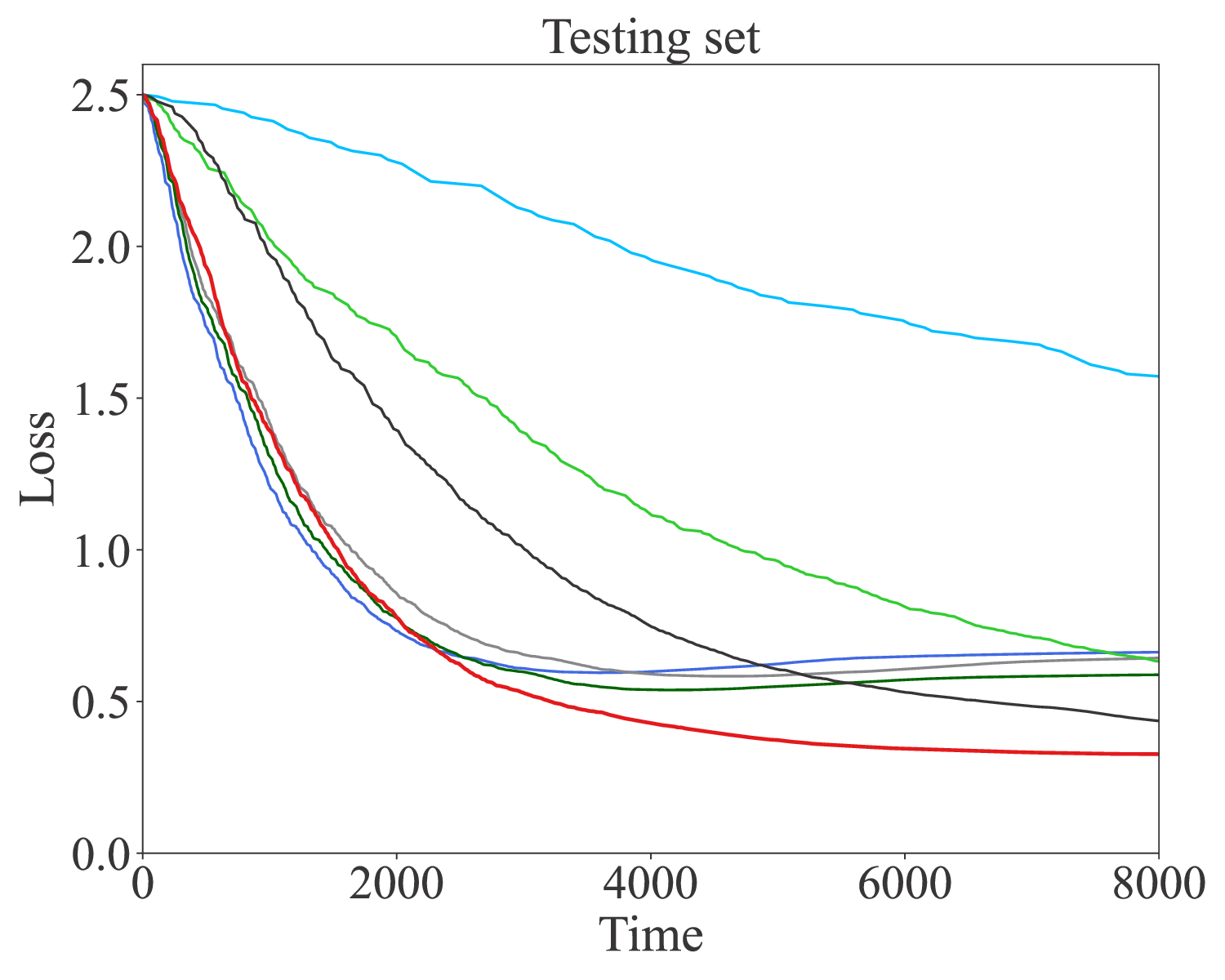}
    \caption{CIFAR10-VGG16}
    \label{fig:cifar10-vgg16-time}
  \end{subfigure}
  \begin{subfigure}{0.325\linewidth}
    \includegraphics[width=1\columnwidth]{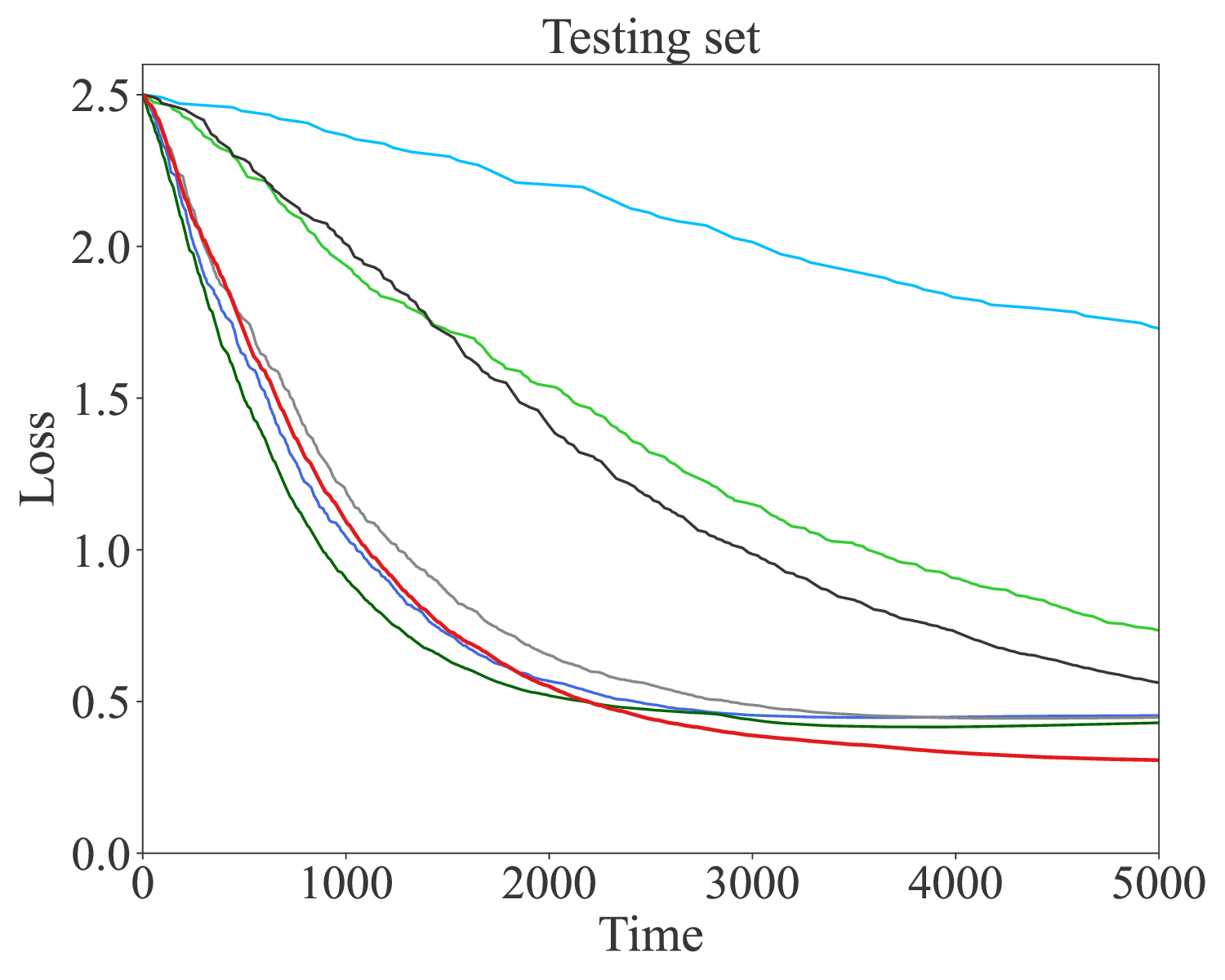}
    \caption{CIFAR10-ResNet18}
    \label{fig:cifar10-resnet18}
  \end{subfigure}
  \begin{subfigure}{0.325\linewidth}
    \includegraphics[width=1\columnwidth]{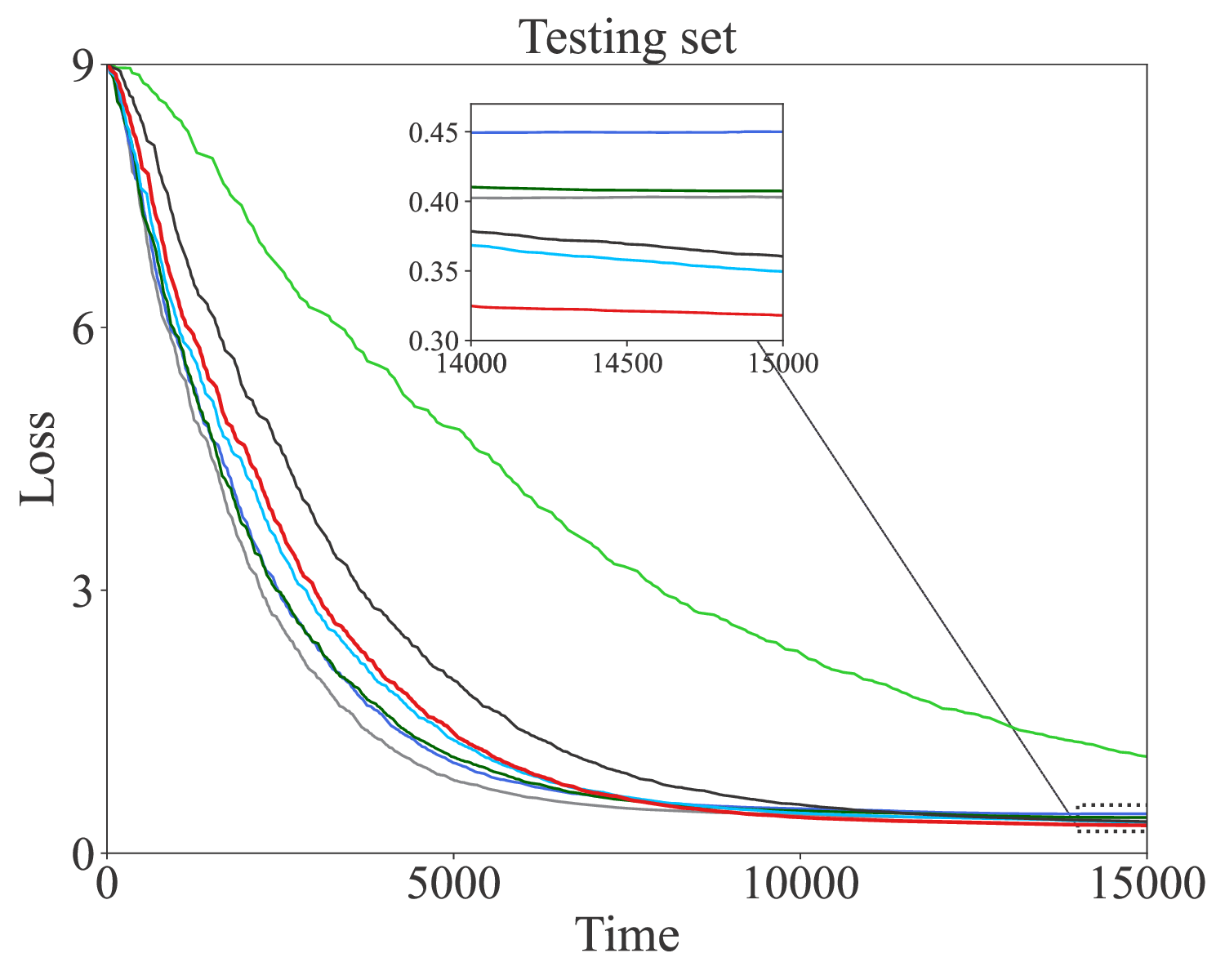}
    \caption{CIFAR10-ResNet50}
    \label{fig:cifar10-resnet50}
  \end{subfigure}
  \begin{subfigure}{0.325\linewidth}
    \includegraphics[width=1\columnwidth]{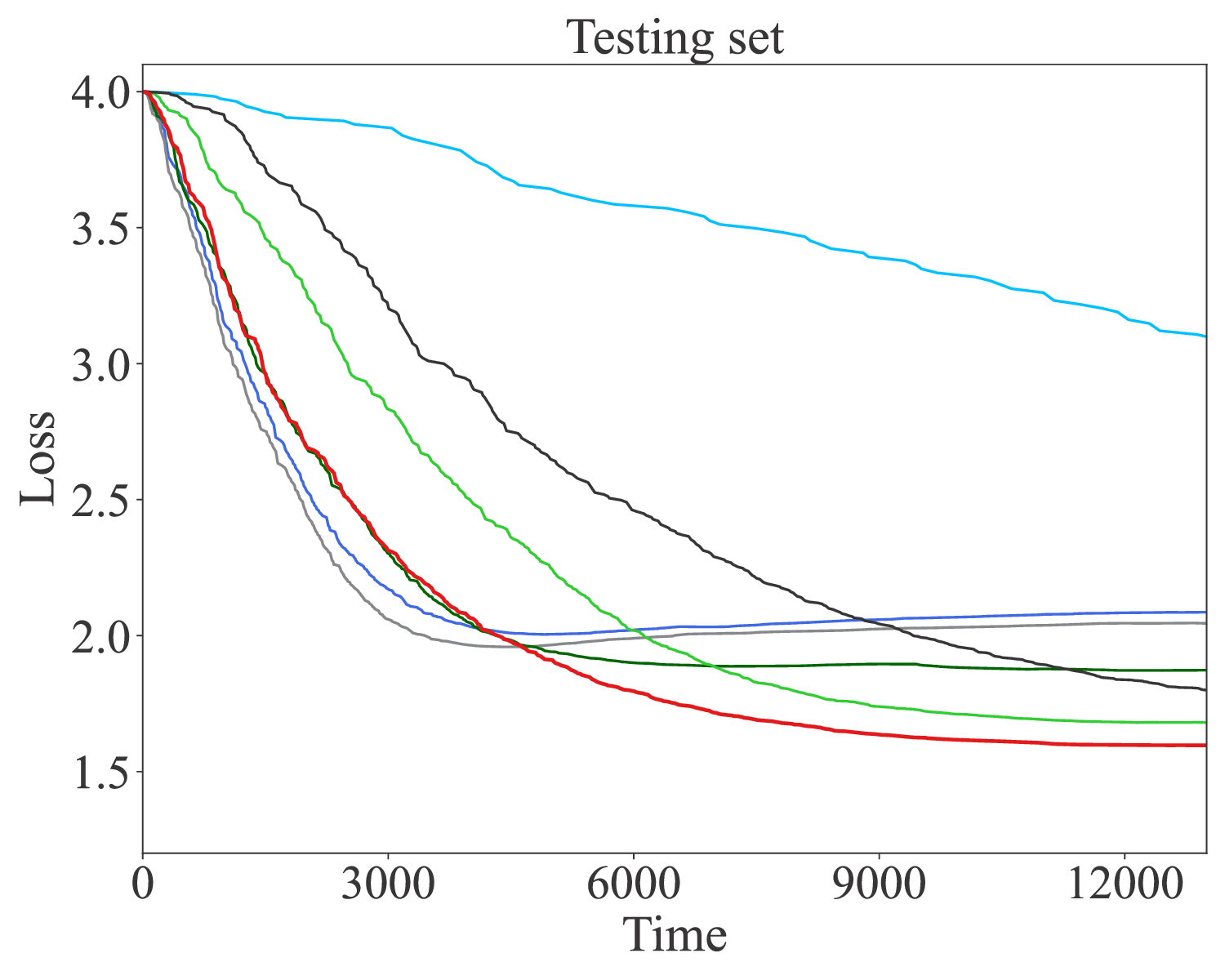}
    \caption{CIFAR100-VGG16}
    \label{fig:cifar100-vgg16-time}
  \end{subfigure}  
  \begin{subfigure}{0.325\linewidth}
    \includegraphics[width=1\columnwidth]{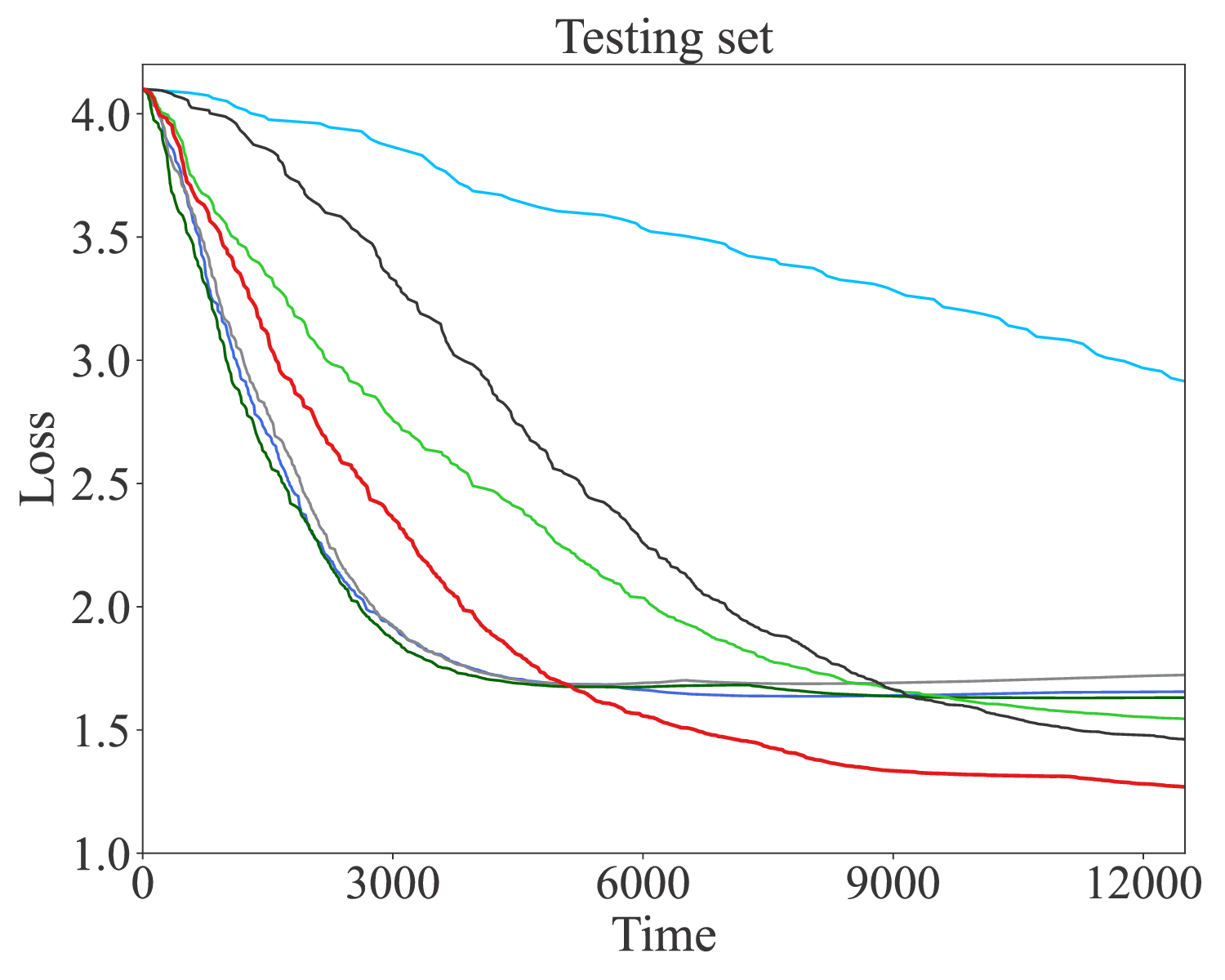}
    \caption{CIFAR100-ResNet18}
    \label{fig:cifar100-resnet18-time}
  \end{subfigure}
    \begin{subfigure}{0.325\linewidth}
    \includegraphics[width=1\columnwidth]{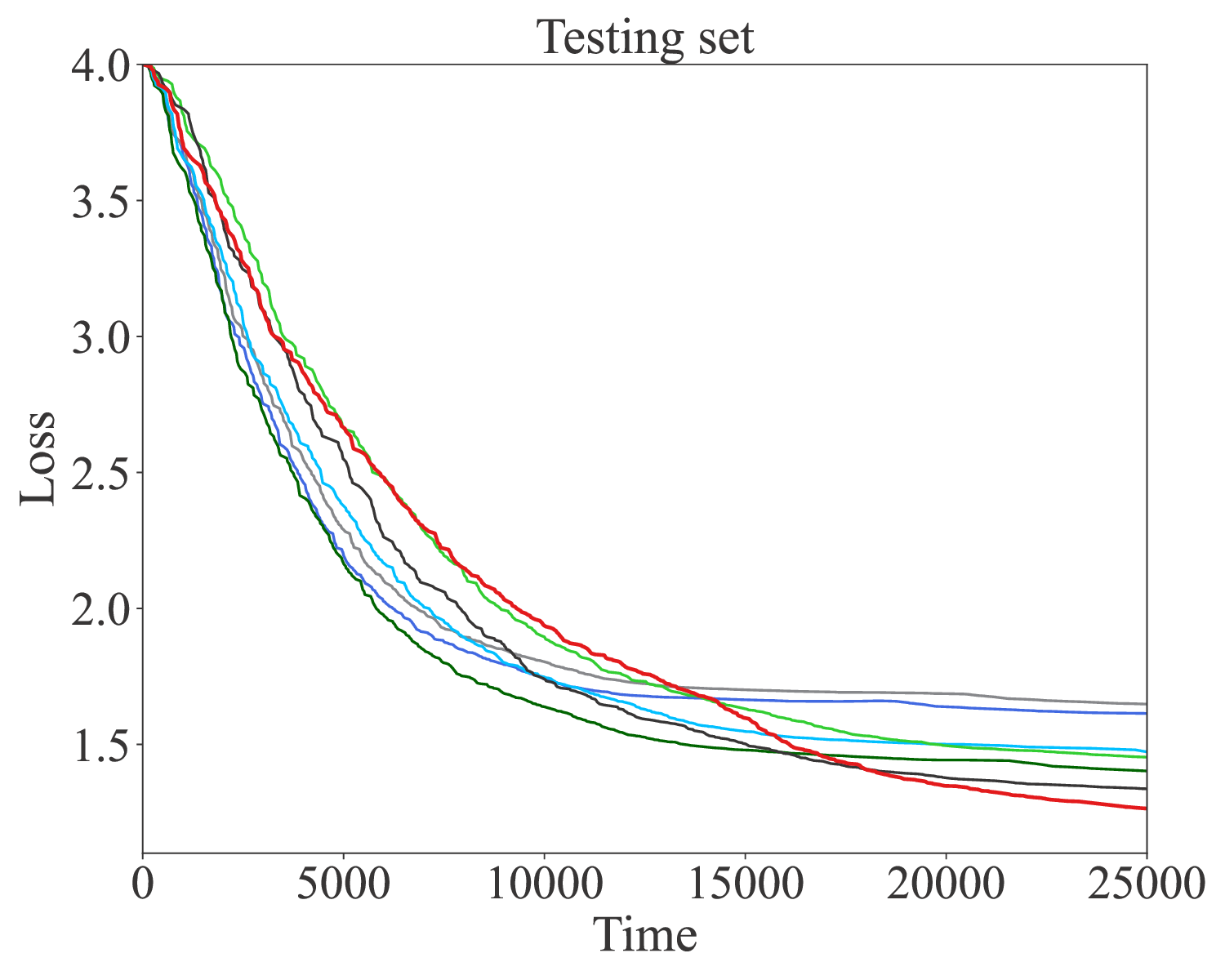}
    \caption{CIFAR100-ResNet50}
    \label{fig:cifar100-resnet50-time}
  \end{subfigure}
    \begin{subfigure}{1\linewidth}
    \includegraphics[width=1\columnwidth]{cifar_legend.eps}
  \end{subfigure}
  \caption{Testing loss vs. wallclock time (seconds) curves of different Riemannian optimization algorithms.}
  \label{loss curve-time}
\end{figure}
\begin{table}[t]
    \centering
    \caption{Memory usage (MB) of optimizer comparisons on optimizing VGG16, ResNet18 and ResNet50. }
    \scalebox{0.7}{
    \begin{tabular}{lccc}
        \toprule
         Methods & VGG16 & ResNet18 & ResNet50\\
         \hline
     	\specialrule{0em} {1.5pt}{1.5pt}
         gmLSTM\citep{9925104} & $2.2849\times 10^{4}$ & $2.2889\times 10^{4}$ &$2.4927\times 10^{4}$\\
         \hline
     	\specialrule{0em} {1.5pt}{1.5pt}
         Ours & $\mathbf{3.9833\times 10^{-2}}$ & $\mathbf{3.9833\times 10^{-2}}$ & $\mathbf{3.9833\times 10^{-2}}$\\
         \bottomrule
    \end{tabular}}

    \label{tab:memory}
\end{table}

\begin{figure}[t]
  \centering
  \begin{subfigure}{0.455\linewidth}
    \includegraphics[width=1.02\columnwidth]{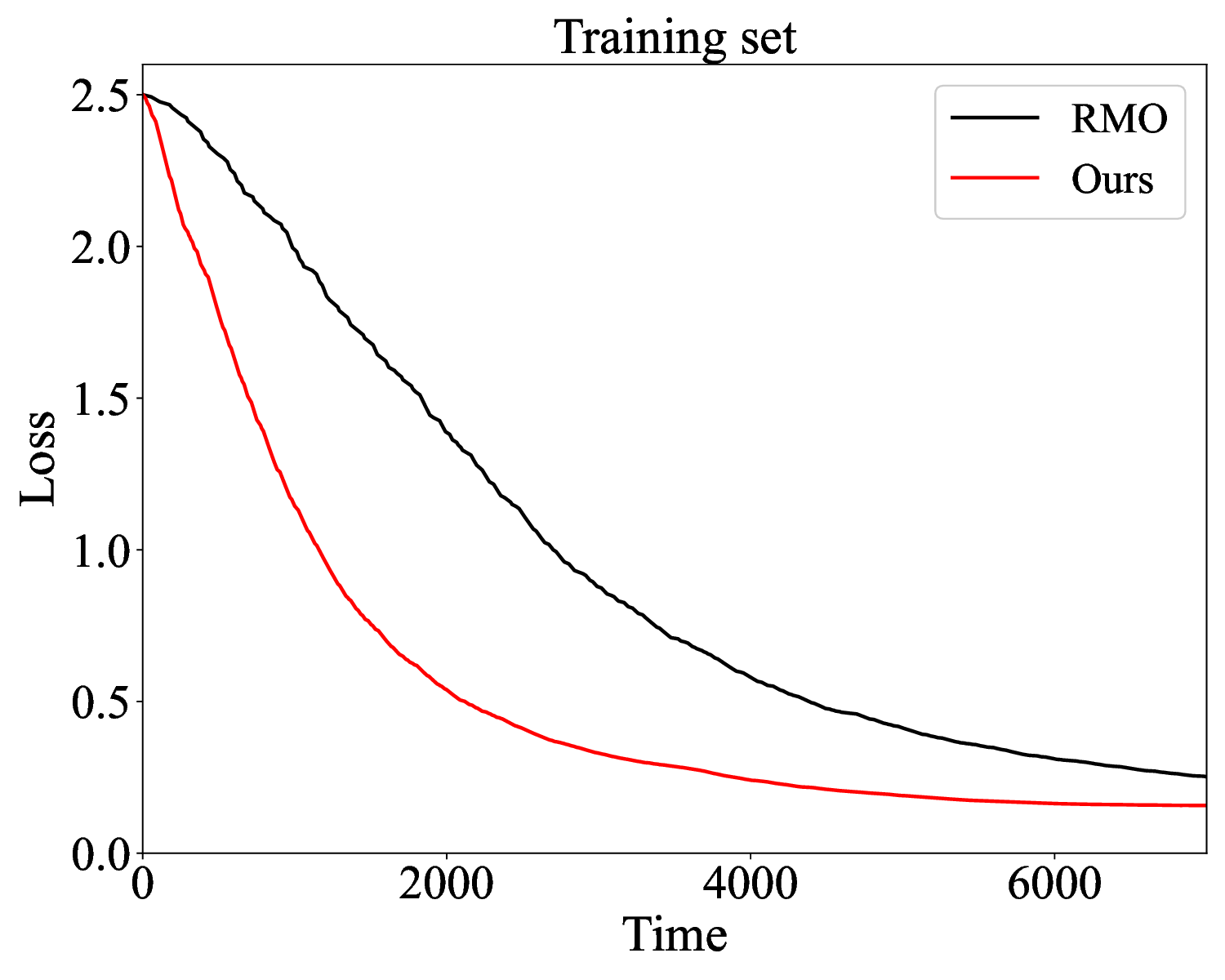}
    \caption{CIFAR10-ResNet18}
    \label{fig:train-a}
  \end{subfigure}
  \hfill
  \begin{subfigure}{0.455\linewidth}
    \includegraphics[width=1.02\columnwidth]{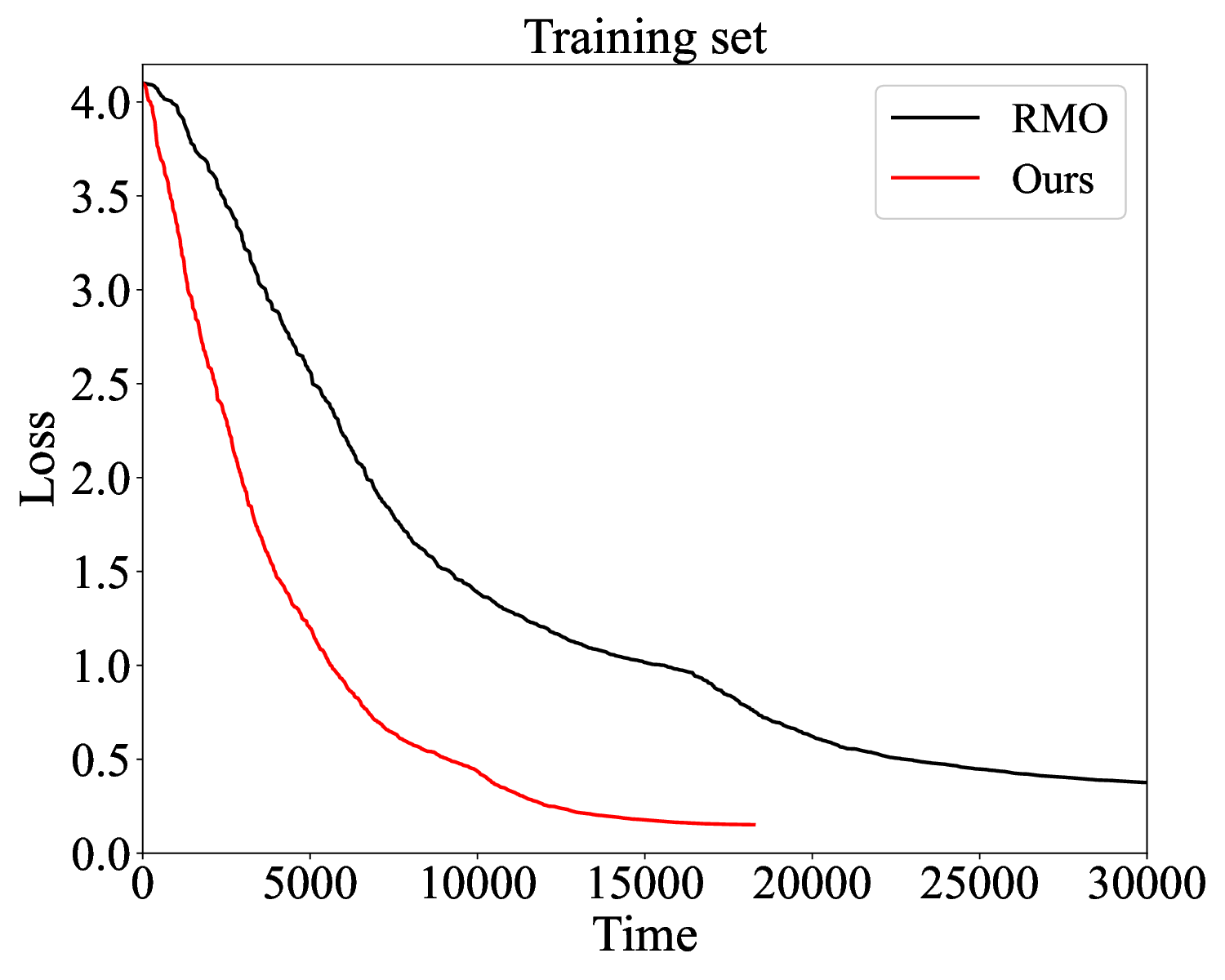}
    \caption{CIFAR100-ResNet18}
    \label{fig:train-b}
  \end{subfigure}
  \caption{{Training loss vs. wallclock time (seconds) curves of optimizing orthogonal ResNet18 using RMO and our method.}}
  \label{training}
\end{figure}

We conduct few-shot image classification experiments on the mini-ImageNet\citep{Vinyals2016MatchingNF} dataset using the ConvNet\citep{Vinyals2016MatchingNF} model. The mini-ImageNet dataset consists of 100 classes from ImageNet\citep{b54} with 600 images for each class, involving 64 training classes, 16 validation classes, and 20 test classes.
Following the standard experimental protocol processed by\citep{Vinyals2016MatchingNF}, we implement 1-shot/5-shot 5 way classification in our experiments. We compare our method with existing few-shot learning optimization based methods: MAML\citep{Finn2017ModelAgnosticMF}, FOMAML\citep{Finn2017ModelAgnosticMF}, Reptile\citep{Nichol2018ReptileAS} and Meta-LSTM\citep{Ravi2017OptimizationAA}, and metric based few-shot learning approaches: Matching Net\citep{Vinyals2016MatchingNF} and ProtoNet\citep{snell2017prototypical}. We also compare our method with the handcrafted Riemannian optimizers. {We set the batchsize $n=256$, the maximum optimization step $\tau=200$ and the maximum inner iteration $T=4$}\par
\begin{table*}[t]

	\centering
	\caption{Classification accuracies(\%) of VGG16, ResNet18 and ResNet50 on the CIFAR10, CIFAR100 and SVHN datasets.}
	\label{resnetvgg}
	\scalebox{0.75}{
	\begin{tabular}{llccccccccc}
		\toprule
            & & &VGG16& & &ResNet18& & &ResNet50&\\
            & Methods& \makecell[c]{CIFAR10}&\makecell[c]{CIFAR100}&\makecell[c]{SVHN}& \makecell[c]{CIFAR10}&\makecell[c]{CIFAR100}&\makecell[c]{SVHN} &\makecell[c]{CIFAR10}&\makecell[c]{CIFAR100}&\makecell[c]{SVHN}\\
		\hline
		\specialrule{0em} {1.5pt}{1.5pt} 
		\makecell[l]{Orthogonalization\\Algorithm}& \makecell[l] {SO\citep{Xie2017AllYN}\\DSO\citep{Bansal2018CanWG}\\ MC\citep{Bansal2018CanWG}\\SRIP\citep{Bansal2018CanWG}\\ONI\citep{Huang2020ControllableOI}} & \makecell[c]{87.87 \\88.40 \\88.15 \\88.08 \\84.10}& \makecell[c]{57.51 \\57.85 \\57.18 \\58.58 \\62.36}& \makecell[c]{93.61 \\95.13 \\94.82 \\95.05 \\93.60}& \makecell[c]{86.88 \\87.01 \\87.10 \\87.51 \\85.02}& \makecell[c]{68.23 \\68.55 \\68.43 \\68.80 \\68.32}& \makecell[c]{94.84 \\94.96 \\94.97 \\95.13 \\95.02}& \makecell[c]{89.64 \\89.70 \\89.53 \\89.67 \\86.24}& \makecell[c]{65.32 \\65.80 \\66.01 \\65.92 \\64.17}& \makecell[c]{91.93 \\92.06 \\92.30 \\92.47 \\91.68}\\
		\hline
		\specialrule{0em} {1.5pt}{1.5pt}
            \makecell[l]{Handcrafted\\Riemannian\\Optimizer}&\makecell[l]{RSGD\citep{Bonnabel2013StochasticGD}\\RSGDM\citep{Roy2018GeometryAC}\\cRMSProp\citep{Roy2018GeometryAC}\\RSVRG\citep{Zhang2016RiemannianSF}\\RASA\citep{Kasai2019RiemannianAS}}&\makecell[c]{87.09 \\87.24 \\89.90 \\89.89 \\89.51}&\makecell[c]{57.39 \\58.40 \\62.16 \\62.37 \\61.59}&\makecell[c]{95.15 \\94.72 \\95.88 \\95.26 \\95.80}&\makecell[c]{87.63 \\87.59 \\89.95 \\89.88 \\89.20}&\makecell[c]{68.12 \\68.92 \\70.79 \\72.15 \\72.35}&\makecell[c]{94.29 \\94.52 \\95.71 \\95.56 \\95.97}&\makecell[c]{90.51 \\91.06 \\91.74 \\91.94 \\91.02}&\makecell[c]{67.49 \\66.83 \\70.94 \\71.06 \\72.30}&\makecell[c]{92.41 \\93.85 \\93.42 \\95.80 \\94.67}\\
		  \hline
		\specialrule{0em} {1.5pt}{1.5pt}
            \makecell[l]{Riemannian\\Meta-optimization}&\makecell[l]{RMO\citep{9925104}\\ Ours}&\makecell[c]{90.30\\\textbf{91.03}}&\makecell[c]{65.40\\\textbf{66.02}}&\makecell[c]{95.92\\\textbf{96.05}}&\makecell[c]{91.02\\\textbf{92.40}}&\makecell[c]{71.31\\\textbf{73.94}}&\makecell[c]{96.10\\\textbf{96.50}}&\makecell[c]{92.43\\\textbf{93.48}}&\makecell[c]{73.66\\\textbf{76.28}}&\makecell[c]{\textbf{96.49}\\95.98}\\

		\bottomrule
	\end{tabular}
	}
\end{table*}

{We also conduct class-incremental learning experiments on the Tiny-ImageNet\citep{le2015tiny} dataset. The class-incremental learning task is ideal for testing the adaptability of our method, as it requires the optimizer to dynamically adapt to new tasks while maintaining performance on previously learned tasks. This aligns with the objectives of our Riemannian meta-optimization approach, which is designed to automatically adapt to dynamic environments. The Tiny-ImageNet dataset contains 200 classes, and each class contains 500 training images, 50 validation images and 50 test images. It has relatively obvious hyperbolic structures and is commonly used in class-incremental learning. We split the Tiny-ImageNet dataset in three different settings: T200-B100-S5, T200-B100-S10 and T200-B100-S20. For instance, T200-B100-S5 represents that we first pre-train the model using the first 100 classes and the following classes are split into 5 stages that each has 20 classes. Deep features that we used are extracted from the convolutional layers of a finetuned ResNet18 model and projected into the hyperbolic space following the work\citep{gao2022hyperbolic}. We project the weight of classifier onto the hyperbolic manifold and use our optimizer to optimize the hyperbolic classifier. We compare our method with Riemannian optimizers: RSGD\citep{Bonnabel2013StochasticGD}, RSGDM, cRMSProp\citep{Roy2018GeometryAC}, RSVRG\citep{Zhang2016RiemannianSF} and RASA\citep{Kasai2019RiemannianAS}. We also compare our method with popular class-incremental learning methods: LwF-MC\citep{rebuffi2017icarl}, iCaRL\citep{rebuffi2017icarl}, MAS\citep{aljundi2018memory} and MUC\citep{liu2020more}. We set $n=16$, $\tau=60$ and $T=8$. We set the curvature as $c=-1.0$. The learning rates and momentum-related hyperparameters of hand-crafted Riemannian optimizers are the same as those in \ref{effec}}

\subsubsection{Memory Cost}
\label{Memory cost}
We analyze the memory cost of using gmLSTM models(\emph{i.e.}, the RMO and I-RMO methods) and our model to optimize the orthogonal neural networks for the general image classification experiments using VGG16, ResNet18 and ResNet50. Unlike our optimizer that can be shared across different Riemannian parameters, the optimizer composed of gmLSTMs are tied to the size of the Riemannian parameters. Thus, we are supposed to deploy corresponding gmLSTMs to learn optimizers for Riemannian parameters with different sizes, making
the memory consumption increase with the size of the optimization problem. 
Results are shown in \autoref{tab:memory}. We observe that deploying gmLSTMs in optimizing orthogonal DNNs on resource-limited devices is impractical due to their extensive storage requirements. In contrast, the memory cost of our model is constant and quite small. The reason is as follows.

For a Riemannian parameter with the size of $d\times p$, the model parameter number of a gmLSTM is $34d^{2} + 1024(dp+1)$. In VGG16, the Riemannian weight matrices $\boldsymbol{W}$ are in sizes of: (576, 64), (576, 128), (1152, 128), (1152, 256), (2304, 256), (2304, 512) and (4608, 512). In this situation, we are supposed to employ 7 gmLSTMs to optimize these Riemannian parameters respectively. The total number of model parameters of these gmLSTMs is about $5.9898 \times 10^9$. Considering each parameter occupies 4 bytes, the memory cost of storing model parameters of these gmLSTMs is $2.2849\times 10^{4}$ MB. In Resnet18, the Riemannian weight matrices $\boldsymbol{W}$ are in sizes of: (147, 64), (576, 64), (576, 128), (1152, 128), (1152, 256), (2304, 256), (2304, 512) and (4608, 512). We are supposed to employ 8 gmLSTMs to optimize these Riemannian parameters respectively. The total number of model parameters of these gmLSTMs is about $6.0002 \times 10^9$. The memory cost of storing model parameters of these gmLSTMs is $2.2889\times 10^{4}$ MB.  
In Resnet50, the Riemannian weight matrices 
$\boldsymbol{W}$ are in sizes of: (576,64), (256, 64), (256, 128), (1152, 128), (512, 128), (512, 256), (2304, 256), (1024, 256), (1024, 512), (4608, 512) and (2048, 512). We are supposed to employ 11 gmLSTMs to optimize these Riemannian parameters respectively. The total number of model parameters of these gmLSTMs is about $6.5344 \times 10^9$. The memory cost of storing model parameters of these gmLSTMs is $2.4927\times 10^{4}$ MB.



In our method, we utilizes two two-layer LSTMs with hidden sizes as 20, and two projection(linear) layers with the size of (20, 1) as the meta model to learn the optimizer. The total parameter number of our model is 10442, and the memory cost is 41768 bytes, that is 40.79 KB. Since the proposed subspace adaptation scheme allows the learned optimizer to be shared across different Riemannian parameters, the memory costs of our model for optimizing the orthogonal VGG16, ResNet18 and ResNet50 are the same and as small as 40.79 KB.

\subsubsection{Convergence Analysis}
We analyze the convergence of the learned
optimizer on the CIFAR10, CIFAR100 and SVHN datasets. Here we do not compare our method with the I-RMO method although it performs well on small-scale optimization problems as shown in \autoref{fig:pca_face} and \autoref{training time}, because I-RMO is difficult to converge on large-scale optimization problems. We analyze the reason is that I-RMO requires to obtain the optimal neural network to derive the outer gradient, while one only can obtain the approximated neural network since the limitation of memory forces I-RMO to set a smaller number of iterations in the inner-loop. Accumulated approximate errors of a large number of parameters in the neural network become huge, hindering the convergence of I-RMO. We compare our learned Riemannian optimizer with the optimizer learned by RMO and the handcrafted Riemannian optimizers.

\autoref{loss curve resnet} shows the loss curves of the learned optimizers of our method and RMO, and hand-designed Riemannian optimizers on the testing set of CIFAR10, CIFAR100 and SVHN. \autoref{loss curve-time} shows the testing loss vs. wallclock time curves on the testing set of CIFAR10 and CIFAR100. We observe that our optimizer converges faster than Riemannian optimizer learned by RMO, and achieves better optima than RMO and hand-crafted optimizers. The results show that our learned optimizer can exploit the underlying data distribution and learn a better optimization trajectory in a data-driven fashion. {\autoref{training} presents the training loss vs. wall-clock time curves. We observe that our optimizer requires less training time than RMO, with the same number of training iterations across both datasets. Moreover, our learned optimizer converges faster and reaches better optima compared to RMO. These results demonstrate that the proposed subspace adaptation scheme significantly reduces training time and accelerates convergence compared to the full-matrix adaptation method.}


\begin{table}[t]
	\normalsize

	\centering
	\caption{Classification accuracies(\%) on the mini-ImageNet dataset. The reported accuracies are with 95\% confidence intervals.}
	\scalebox{0.65}{
	\begin{tabular}{llc}
		\toprule
		 & Methods & \makecell[c]{Accuracy$(\%)$ \\ 5-way-1-shot\hspace{0.2cm} 5-way-5-shot}\\
		\hline
		\specialrule{0em} {1.5pt}{1.5pt}
		\makecell[l]{Few-shot\\ Learning\\ Algorithm} & \makecell[l]{Matching Net\citep{Vinyals2016MatchingNF}\\ProtoNet\citep{snell2017prototypical}\\meta-LSTM\citep{Ravi2017OptimizationAA}\\MAML\citep{Finn2017ModelAgnosticMF}\\FOMAML\citep{Finn2017ModelAgnosticMF}\\Reptile\citep{Nichol2018ReptileAS}} & \makecell[c]{43.56 $\pm$ 0.84\hspace{0.32cm} 55.31 $\pm$ 0.73\\44.42 $\pm$ 0.84\hspace{0.32cm} 64.24 $\pm$ 0.72\\ 43.33 $\pm$ 0.77\hspace{0.32cm} 60.60 $\pm$ 0.71\\46.21 $\pm$ 1.76\hspace{0.32cm} 61.12 $\pm$ 1.01\\ 45.53 $\pm$ 1.58\hspace{0.32cm} 62.02 $\pm$ 1.12\\47.07 $\pm$ 0.26\hspace{0.32cm} 62.74 $\pm$ 0.33}\\
		\hline
		\specialrule{0em} {1.5pt}{1.5pt}
		\makecell[l]{Handcrafted\\Riemannian\\Optimizer}& \makecell[l]{RSGD\citep{Bonnabel2013StochasticGD} \\ RSGDM\citep{Roy2018GeometryAC} \\ RSVRG\citep{Zhang2016RiemannianSF} \\ cRMSProp\citep{Roy2018GeometryAC} \\ RASA\citep{Kasai2019RiemannianAS}}&  \makecell[c]{43.63 $\pm$ 1.26\hspace{0.32cm} 60.15 $\pm$ 1.15\\44.02 $\pm$ 1.39\hspace{0.32cm} 60.76 $\pm$ 1.18\\45.93 $\pm$ 1.33\hspace{0.32cm} 62.55 $\pm$ 1.24\\46.01 $\pm$ 1.35\hspace{0.32cm} 62.83 $\pm$ 1.29\\44.29 $\pm$ 1.43\hspace{0.32cm} 61.38 $\pm$ 1.20}\\
		\hline
     	\specialrule{0em} {1.5pt}{1.5pt}
		\makecell[l]{Riemannian\\Meta-optimization}&\makecell[l]{Ours} & \makecell[c]{48.72 $\pm$ 1.56 \hspace{0.32cm} \textbf{65.87 $\pm$ 1.31} }
		\\
		\bottomrule
	\end{tabular}}
	\label{mini}
\end{table}

\begin{table*}[t]
\caption{{Final accuracy (\%) on the Tiny-ImageNet dataset with different numbers of incremental stages.}}
\centering
\arrayrulewidth=1pt 

\begingroup
\scalebox{0.8}{
\begin{tabular}{llccc}
		\toprule
             & &Tiny-ImageNet& & \\
            & Methods& \makecell[c]{T200-B100-S5}&\makecell[c]{T200-B100-S10}&\makecell[c]{T200-B100-S20}\\
		\hline
		\specialrule{0em} {1.5pt}{1.5pt} 
		\makecell[l]{Class-incremental\\Learning Algorithm}& \makecell[l] {MAS\citep{aljundi2018memory}\\LwF-MC\citep{rebuffi2017icarl}\\MUC\citep{liu2020more}\\iCaRLCNN\citep{rebuffi2017icarl}} & \makecell[c]{11.64 \\15.56 \\17.23 \\23.17}& \makecell[c]{6.56 \\13.26 \\15.21 \\20.71  }& \makecell[c]{3.23 \\7.87 \\8.26 \\20.28 }\\
		\hline
		\specialrule{0em} {1.5pt}{1.5pt}
            \makecell[l]{Handcrafted\\Riemannian\\Optimizer}&\makecell[l]{RSGD\citep{Bonnabel2013StochasticGD}\\RSGDM\citep{Roy2018GeometryAC}\\cRMSProp\citep{Roy2018GeometryAC}\\RSVRG\citep{Zhang2016RiemannianSF}\\RASA\citep{Kasai2019RiemannianAS}}&\makecell[c]{14.88 \\15.87 \\26.03\\31.60  \\31.98}&\makecell[c]{12.70 \\ 13.45 \\23.28 \\30.85\\31.05}&\makecell[c]{10.95 \\11.27  \\21.86 \\29.84\\30.27}\\
		  \hline
		\specialrule{0em} {1.5pt}{1.5pt}
            \makecell[l]{Riemannian\\Meta-optimization}&\makecell[l]{Ours}&\makecell[c]{\textbf{32.44}}&\makecell[c]{\textbf{31.52}}&\makecell[c]{\textbf{30.92}}\\
		\bottomrule
	\end{tabular}}
\endgroup

\label{table:tinyimagenet}
\end{table*}

\subsubsection{Accuracy Analysis}
We report the accuracies of our method on the general image classification tasks and few-shot image classification task. \autoref{resnetvgg} shows classification accuracy results of VGG16, ResNet18 and ResNet50 on the CIFAR10, CIFAR100 and SVHN datasets. As shown in the table, the proposed method clearly outperforms the best in training of all three networks on the CIFAR10 and CIFAR100 datasets. On the SVHN dataset, our method achieves comparable performance with RMO and better results than other compared algorithms.

\autoref{mini} reports the classification accuracy results on the few-shot learning task using the mini-ImageNet dataset. The results show that our method achieves comparable results and even outperforms other methods considered for comparison for both 1-shot and 5-shot experiments. Compared with the handcrafted Riemannian optimizers, our optimizer learned in a data-driven manner enjoys better generalization performance for new tasks.

{\autoref{table:tinyimagenet} reports the final accuracies for the T200-B100-S5, T200-B100-S10, and T200-B100-S20 settings. We observe that our method significantly outperforms the hand-crafted Riemannian optimizers. This indicates that our learned optimizer can generate suitable step sizes and search directions for Riemannian parameters at different stages of class-incremental learning, which helps mitigate catastrophic forgetting to a certain extent. When compared with representative class-incremental learning methods, our method performs competitively or even surpasses them in some settings, demonstrating the effectiveness of our method.}

\begin{figure}[t]
  \centering
  \begin{subfigure}{0.455\linewidth}
    \includegraphics[width=1\columnwidth]{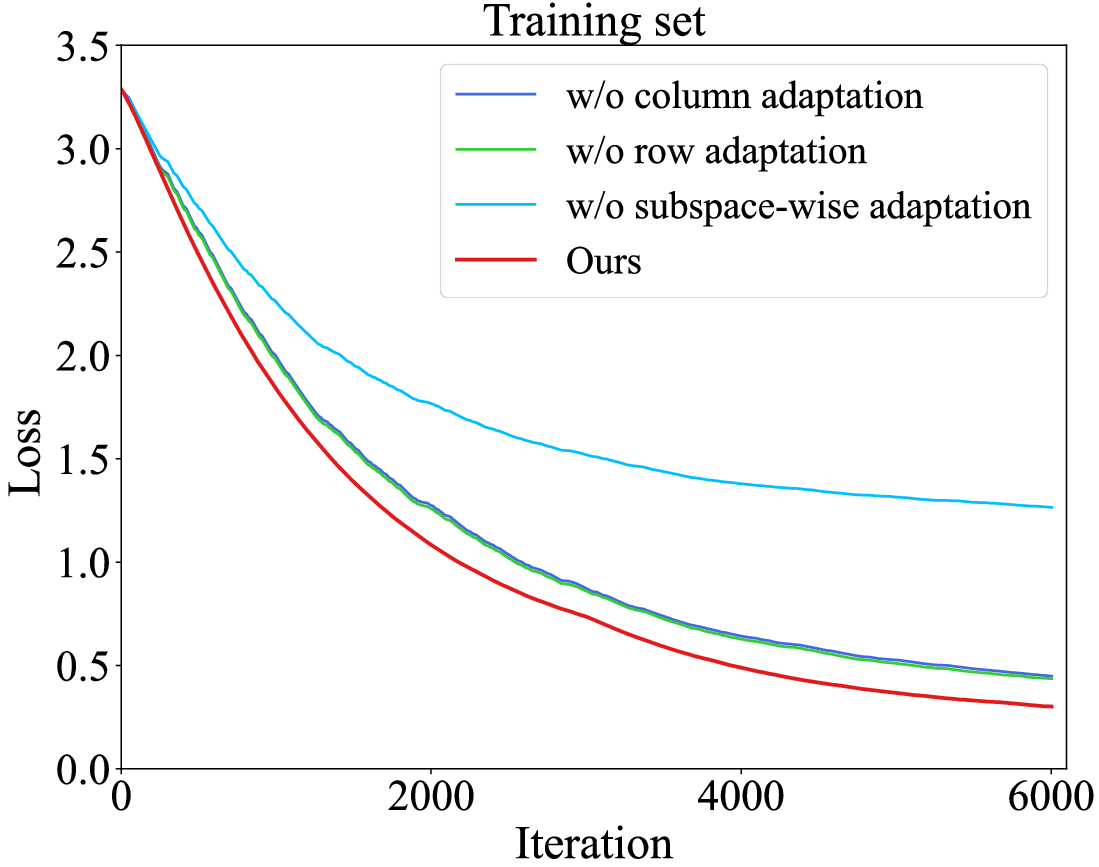}
    \caption{Training set-VGG16}
    \label{fig:ablation-a}
  \end{subfigure}
  \begin{subfigure}{0.455\linewidth}
    \includegraphics[width=1\columnwidth]{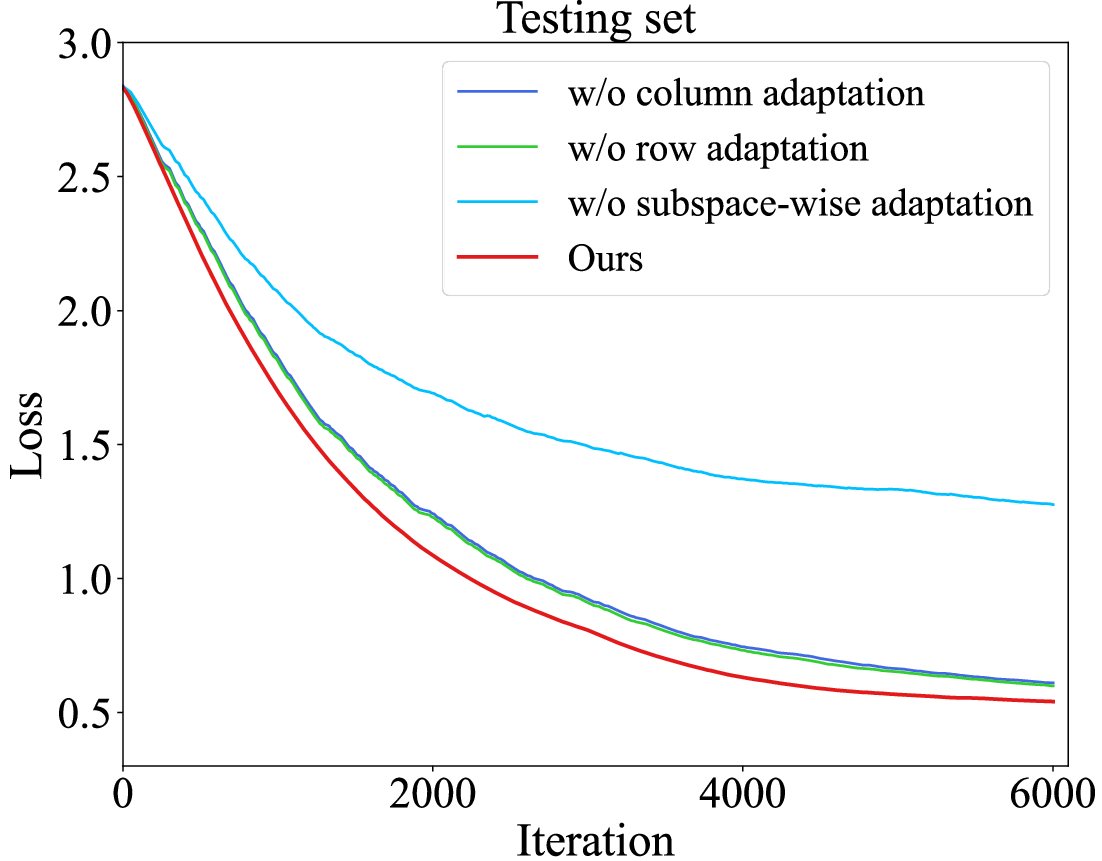}
    \caption{Testing set-VGG16}
    \label{fig:ablation-b}
    \end{subfigure}
        \begin{subfigure}{0.455\linewidth}
    \includegraphics[width=1.02\columnwidth]{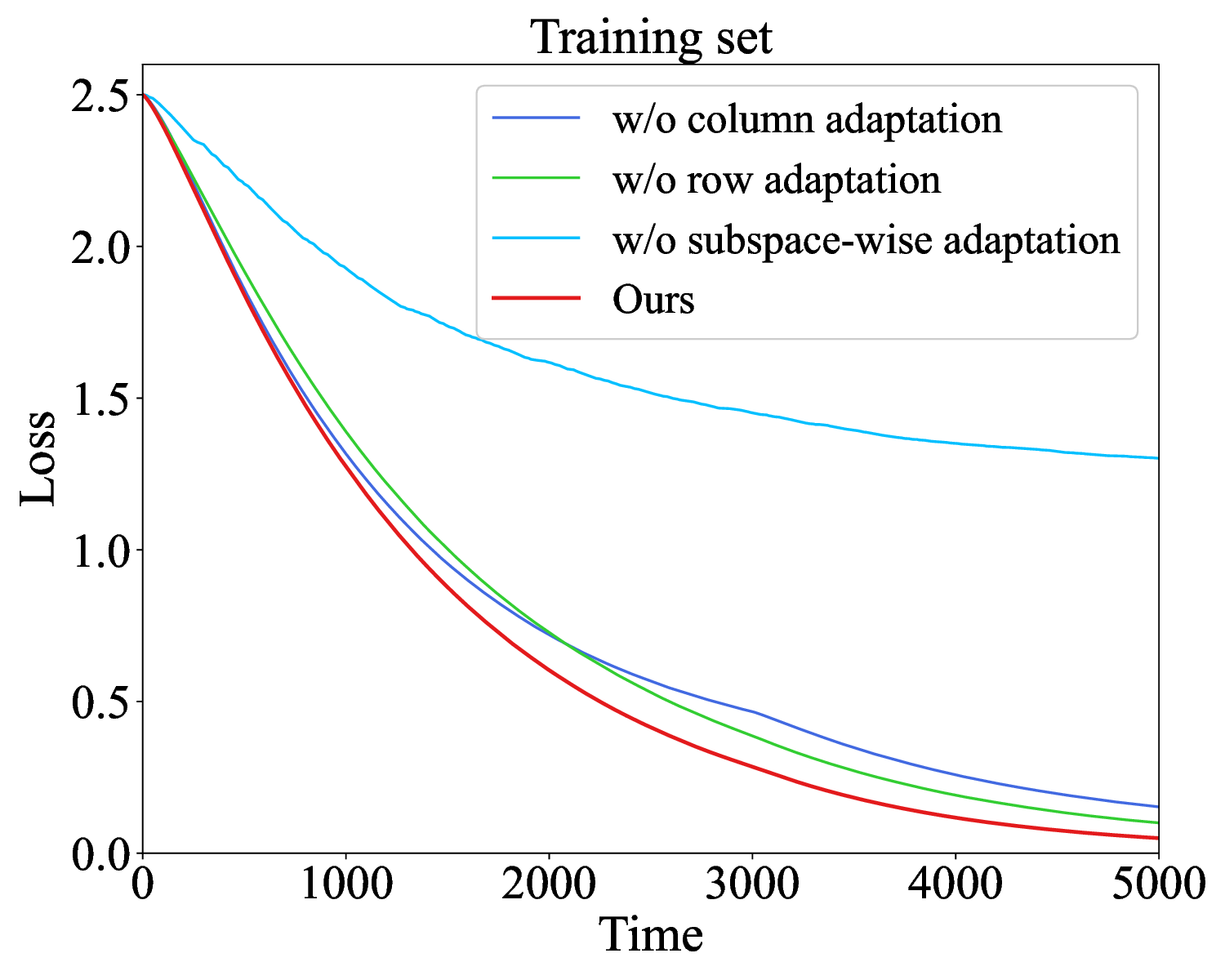}
    \caption{Training set-ResNet18}
    \label{fig:ablation-c}
  \end{subfigure}
  \begin{subfigure}{0.455\linewidth}
    \includegraphics[width=1.02\columnwidth]{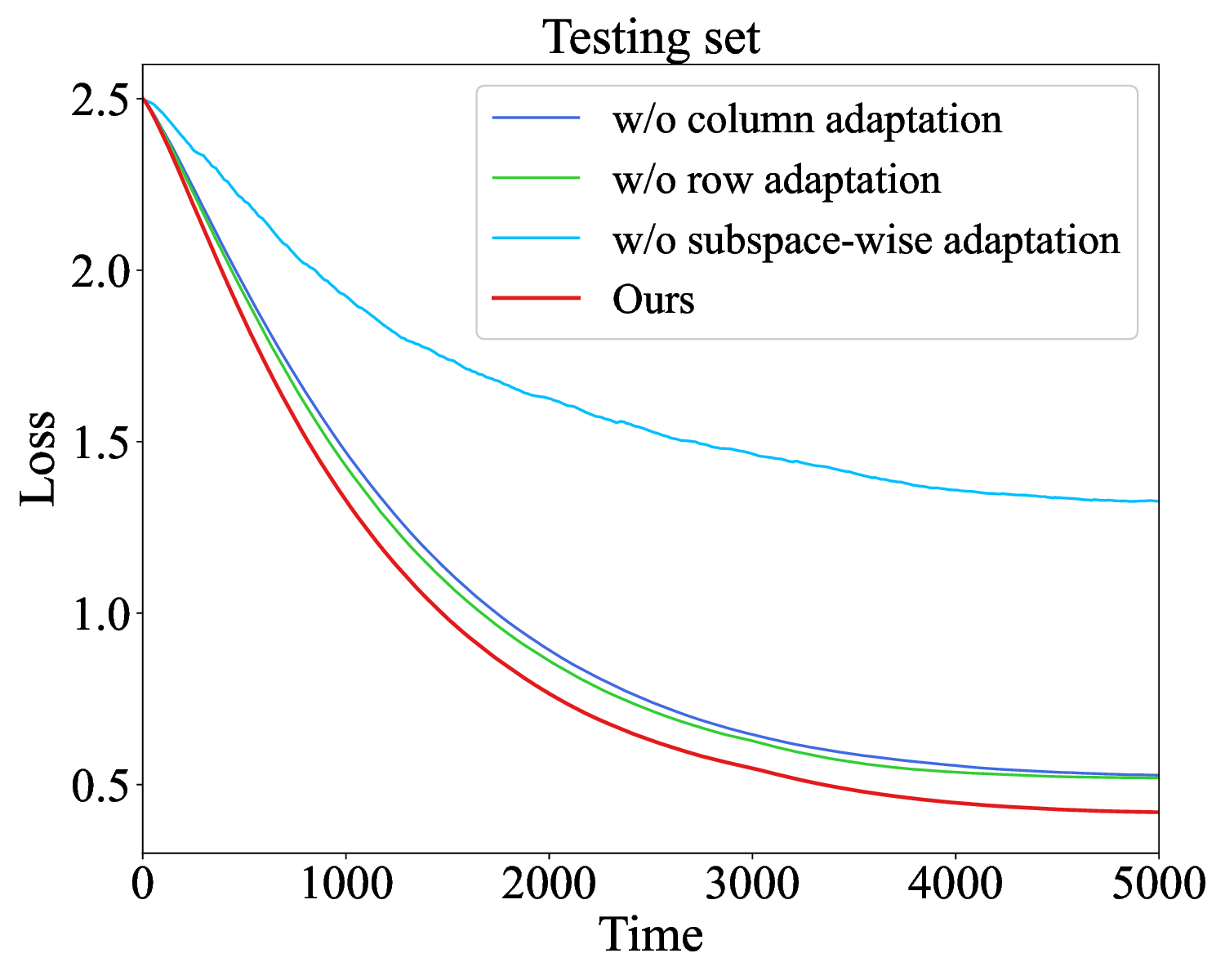}
    \caption{Testing set-ResNet18}
    \label{fig:ablation-d}
    \end{subfigure}
  \caption{Plots for the ablation experiments on the CIFAR10 dataset.}
  \label{ablation}
\end{figure}

\subsection{Ablation Study}
In this section, we conduct ablation experiments in image classification on CIFAR10 using VGG16 and ResNet18 to evaluate the proposed subspace adaptation scheme. Concretely, we compare our method with: (1) we remove the row subspace adaptation and only adapt the column subspace of $\boldsymbol{G}$; (2) we remove the column subspace adaptation and only adapt the row subspace of $\boldsymbol{G}$; (3) we remove the subspace adaptation scheme and directly use conventional LSTM model to produce the step sizes and search directions. We follow the same experimental settings in the general image classification experiments.

\begin{table}[t]
    \centering

    \caption{Classification accuracies(\%) on the CIFAR10 dataset for ablation experiments. }
    \scalebox{0.75}{
    \begin{tabular}{lc}
        \toprule
         Methods & \makecell[c]{Accuracy$(\%)$ \\ VGG16\hspace{0.7cm} ResNet18}  \\
         \hline
     	\specialrule{0em} {1.5pt}{1.5pt}
         w/o subspace adaptation &
         \makecell[c]{78.40 \hspace{0.7cm} 76.83 }\\
     	\specialrule{0em} {1.5pt}{1.5pt}
         row subspace adaptation & \makecell[c]{87.86 \hspace{0.7cm} 87.80} \\
     	\specialrule{0em} {1.5pt}{1.5pt}
         column subspace adaptation & \makecell[c]{87.57 \hspace{0.7cm} 87.05} \\
     	\specialrule{0em} {1.5pt}{1.5pt}
         Ours & \makecell[c]{\textbf{91.03} \hspace{0.7cm} \textbf{92.40} }\\
         \bottomrule
    \end{tabular}}
    \label{tab:abaltion}
\end{table}

\autoref{ablation} shows the loss curves on the training and testing sets of CIFAR10 for optimizing VGG16 and ResNet18. We observe that without subspace adaptation, the optimizer parameterized only by conventional LSTM, or by row or column subspace adaptation converges slowlier. Table~\ref{tab:abaltion} shows classification accuracy results on the CIFAR10 dataset. As shown in the table, the optimizer parameterized by conventional LSTM without subspace adaptation obtains the worst results. The optimizers parameterized only
by row or column subspace adaptation also get worse accuracies than our optimizer on both models. The results demonstrate the effectiveness of the proposed subspace adaptation scheme.


\section{Conclusion}
In this paper, we have presented an efficient Riemannian meta-optimization method that uses the subspace adaptation scheme to reduce the heavy memory footprint in large-scale optimization settings. The proposed subspace adaptation scheme can avoid directly adapting the whole Riemannian gradients and allow the learned optimizer to be shared across different Riemannian parameters, which significantly reduces the memory loads.
We demonstrate empirically that our method only requires a small constant memory cost to perform effective Riemannian meta-optimization. Compared with existing Riemannian meta-optimization methods on optimizing an orthogonal ResNet18 network, our method significantly reduces the memory consumption from 22.35GB to only 40.79KB.  

{Our approach incorporates meta-learning to train the optimizer, where the learned prior knowledge enables it to dynamically adjust step sizes and search directions across different tasks. This makes our method well-suited for open-environment challenges, such as continual learning and zero-shot recognition, where both the environment and objectives are constantly changing. In such scenarios, our optimizer’s ability to adapt to different tasks
can help mitigate problems like catastrophic forgetting and improve generalization to new classes. We plan to further explore these promising directions in future work. }In addition, although we have demonstrated through experiments the convergence  of the optimizer learned via bi-level optimization and its generalization ability to new tasks, it still lacks rigorous theoretical guarantees.  
In our future work, we will consider bounding the gradients of learned parameters for the optimizer in the outer loop to prove its convergence\citep{franceschi2018bilevel,liu2020generic}, and conduct the generalization analysis by drawing inspiration from the PAC generalization error bound theorem\citep{foret2020sharpness,DziugaiteR17}.

\textbf{Acknowledgements.}
This work was supported by the Key Program of the National Natural Science Foundation of Shenzhen under Grant No. 202412023000734, the Natural Science Foundation of Shenzhen under Grant No. JCYJ20230807142703006, and the Key Research Platforms and Projects of the Guangdong Provincial Department of Education under Grant No.2023ZDZX1034.



\bibliographystyle{model2-names}
\bibliography{refs}

\end{document}